\RequirePackage{letltxmacro}
\LetLtxMacro{\LaTeXtextbf}{\textbf}
\documentclass{ieeeaccess}
\LetLtxMacro{\textbf}{\LaTeXtextbf}

\IEEEoverridecommandlockouts

\usepackage{times}
\usepackage{makecell}
\usepackage{amsmath}
\usepackage{longtable}
\usepackage{amssymb}
\usepackage{array}
\usepackage{dblfloatfix}
\usepackage{booktabs}
\usepackage{bigstrut}
\usepackage{multirow,bigdelim}
\usepackage{multirow}%
\usepackage{amsfonts}
\usepackage{xcolor}
\usepackage{soul}

\usepackage{cite}
\usepackage{amsmath,amssymb,amsfonts}
\usepackage{textcomp}
\usepackage{comment}

\usepackage[titlenumbered,ruled]{algorithm2e}
\usepackage{algpseudocode}
\usepackage{setspace}
\usepackage{helvet}
\usepackage{courier}
\usepackage{MnSymbol}
\usepackage{mathdots}
\usepackage{graphicx}
\usepackage{color}
\usepackage{amsthm}
\usepackage{pdfpages}

\newtheorem{innercustomgeneric}{\customgenericname}
\providecommand{\customgenericname}{}
\newcommand{\newcustomtheorem}[2]{%
  \newenvironment{#1}[1]
  {%
   \renewcommand\customgenericname{#2}%
   \renewcommand\theinnercustomgeneric{##1}%
   \innercustomgeneric
  }
  {\endinnercustomgeneric}
}

\newcustomtheorem{customthm}{Theorem}
\newcustomtheorem{customproposition}{Proposition}

\def\BibTeX{{\rm B\kern-.05em{\sc i\kern-.025em b}\kern-.08em
    T\kern-.1667em\lower.7ex\hbox{E}\kern-.125emX}}
\begin{document}
\history{Date of publication xxxx 00, 0000, date of current version xxxx 00, 0000.}
\doi{10.1109/ACCESS.2017.DOI}

\title{Constrained Clustering: General Pairwise and Cardinality Constraints}
\author{\uppercase{Adel Bibi}\authorrefmark{1}, \uppercase{Ali Alqahtani}\authorrefmark{2,3}, \uppercase{and Bernard Ghanem}\authorrefmark{4}, 
}

\address[1]{Department of Engineering Science, University of Oxford, UK (e-mail: adel.bibi@eng.ox.ac.uk)}

\address[2]{Department of Computer Science, King Khalid University, Abha 61421, Saudi Arabia (e-mail: amosfer@kku.edu.sa)}
\address[3]{Center for Artificial Intelligence (CAI), King Khalid University, Abha 61421, Saudi Arabia}
\address[4]{King Abdullah University of Science and Technology, Thuwal, Saudi Arabia (e-mail: bernard.ghanem@kaust.edu.sa)}

\markboth
{Author \headeretal: Preparation of Papers for IEEE TRANSACTIONS and JOURNALS}
{Author \headeretal: Preparation of Papers for IEEE TRANSACTIONS and JOURNALS}

\corresp{Corresponding author: Adel Bibi}

\begin{abstract}
In this work, we study constrained clustering, where constraints are utilized to guide the clustering process. In existing works, two categories of constraints have been widely explored, namely pairwise and cardinality constraints. Pairwise constraints enforce the cluster labels of two instances to be the same (must-link constraints) or different (cannot-link constraints). Cardinality constraints encourage cluster sizes to satisfy a user-specified distribution. However, most existing constrained clustering models can only utilize one category of constraints at a time. In this paper, we enforce the above two categories into a unified clustering model starting with the integer program formulation of the standard K-means. As these two categories provide useful information at different levels, utilizing both of them is expected to allow for better clustering performance. However, the optimization is difficult due to the binary and quadratic constraints in the proposed unified formulation. To alleviate this difficulty, we utilize two techniques: equivalently replacing the binary constraints by the intersection of two continuous constraints; the other is transforming the quadratic constraints into bi-linear constraints by introducing extra variables. 
Then we derive an equivalent continuous reformulation with simple constraints, which can be efficiently solved by Alternating Direction Method of Multipliers (ADMM) algorithm. Extensive experiments on both synthetic and real data demonstrate: \textbf{(1)} when utilizing a single category of constraint, the proposed model is superior to or competitive with state-of-the-art constrained clustering models, and \textbf{(2)} when utilizing both categories of constraints jointly, the proposed model shows better performance than the case of the single category. The experimental results show that the proposed method exploits the constraints to achieve perfect clustering performance with improved clustering to $2-5$\% in classical clustering metrics, e.g. Adjusted Random Index (ARI), Mirkin's Index (MI), and Huber's Index (HI), outerperfomring all compared-againts methods across the board. Moreover, we show that our method is robust to initialization.

\begin{keywords}
Constrained Clustering, K-means, Pairwise, Cardinality Constraints
\end{keywords}
\end{abstract}

\titlepgskip=-15pt

\maketitle

\section{Introduction}
Clustering is the task of partitioning data into different clusters, based on some specific cluster assumptions. For example, K-means and Gaussian mixture models (GMM) assume each cluster is sampled from a Gaussian distribution. In contrast,
density-based clustering assumes that the densities of data points in different clusters should be different, such as Chameleon \cite{karypis1999chameleon} and AITC \cite{wu2011density}, or clusters should be partitioned at low density regions
\cite{chapelle2005semi}. However, if the adopted cluster assumption is not suited to the target dataset, this may result in a poor performance. To avoid such performance instability, prior knowledge or constraints on the data can be used to guide
the clustering process. These constraints are independent of cluster assumptions, and they provide weak supervision to reflect user preferences. Thus, clustering with constraints, called {\it constrained clustering},
\cite{wagstaff2000clustering,klein2002instance,ng2002spectral}, is expected to give better and more stable performance than unconstrained clustering.

Two main categories of constraints have been widely studied in the field of constrained clustering, namely pairwise and cardinality constraints. \textit{Pairwise constraints} may arise from some form of perceived similarity between samples. For
instance, the continuity property is a form of \textit{Pairwise constraints} that suggests that neighbouring samples are likely to be clustered together and vice versa. Thus, \textit{Pairwise constraints} include must-link and cannot-link
constraints. Must-link constraints enforce that a set of pairs of instances should be in the same cluster, while cannot-link constraints enforce that they belong to different clusters. Thereafter, this category can be viewed as instance-level
constraints. On the other hand, \textit{Cardinality constraints} provide extra knowledge on the size distribution of all clusters. This sort of constraints become particularly necessary in clustering tasks of data that is high dimensional and
sparse with many clusters to assign \cite{bradley2000constrained}. This often leads to solutions of empty clusters or unbalanced cluster assignments. Balancing constraints that lead to equal sized clusters are only a special case of
\textit{Cardinality constraints}. This category in general can be viewed as cluster-level constraints.

Many clustering methods have been proposed to utilize one of the two categories of constraints, such as the ones with pairwise constraints
\cite{lu2010constrained,von2007tutorial,lu2007penalized,dai2003techniques,wu2013constrained,ng2002spectral,wagstaff2001constrained,klein2002instance,wagstaff2000clustering}, and the ones with cardinality constraints
\cite{hoppner2008clustering,klawonn2006equi,bradley2000constrained,shi2000normalized}. However, in some cases, one might want to enforce the continuity property among a set of points and the same time requiring to have solutions of balanced or user
specified cluster sizes. In general, both constraints can be provided simultaneously, as they are derived from different sources. For example, pairwise constraints are usually obtained from an oracle query, while cardinality constraints can be
obtained from experience or user preference. Moreover, they represent supervision at different levels. Each of them can provide particularly useful information that is not covered by the other. Thus, having both sets of constraints together in a
clustering task should signifecently improve the performance and to the best of our knowledge, there is no existing work that can seamlessly incorporate both categories jointly.

For existing constrained clustering methods that handle one constraint category, it is non-trivial to directly add the other. For example, embedding cardinality constraints into the COP-KMEANS \cite{wagstaff2001constrained}, will lead to
instability in performance where COP-KMEANS will often fail in finding a feasible solution. This is because COP-KMEANS is very sensitive to cluster initialization. Moreover, it is also not easy to embed the pairwise constraints into
normalized/ratio-cut \cite{shi2000normalized}, which exploits balanced distribution constraints. In short, existing models are designed to exploit one category of constraints at a time.

We propose a unified model to incorporate both categories of constraints to guide the clustering process. 
Specifically, we start from the formulation of the standard K-mean method, and formulate cardinality constraints into linear constraints and pairwise constraints into quadratic constraints. Then we obtain a discrete optimization problem with quadratic constraints, which is difficult to be solved by off-the-shelf optimization methods. Thus we propose to utilize two techniques. One is to equivalently replace the binary constraints by the intersection of two continuous constraints, which was firstly proposed in \cite{Wu2016ell_pBoxAA}. The other is to transform the quadratic constraints to bi-linear constraints by introducing extra variables. Our key contributions revolve around the new novel continuious reformulation for the K-Means problem allowing to incorporate both cardinality and pair-wise constraints. The reformulation is simple, flexible, and enjoys nice convergence properties with competitive performance. The contributions can be summarized in three folds.
\begin{itemize}
    \item We embed both pairwise and cardinality constraints into one unified clustering model. To the best of our knowledge, this is the first attempt in the field of constrained clustering.
    \item We propose to equivalently transform the binary and quadratic constraints in the original problem to continuous and bi-linear constraints, to obtain a simple continuous reformulation.
    \item We conduct various experiments on several synthetic and real datasets comparing against 5 different algorithms. Our approach demonstrate competitive edge over all compared-against methods in the final clustering performance while respecting the imposed constraints.
\end{itemize}

\section{Related Work}
Here, we briefly review existing clustering models that utilize pairwise or cardinality constraints.

\noindent\textbf{Pairwise Constraints.} They were first introduced into clustering in \cite{wagstaff2000clustering} and \cite{wagstaff2001constrained}. In \cite{wagstaff2000clustering}, a method called COP-COBWEB inserted the pairwise
constraints into the clustering process of the incremental clustering method COBWEB \cite{COBWEB-1987}, which utilizes four operators (i.e. add, new, merge, and split) to maximize the intra-cluster similarity and the inter-cluster dissimilarity. In
each operator of COP-COBWEB, the given pairwise constraints are checked to ensure the satisfaction of all constraints. In \cite{wagstaff2001constrained}, the method COP-KMEANS checks the pairwise constraints in each assignment step of K-means. Both
COP-COBWEB and COP-KMEANS treat the pairwise constraints as hard constraints (i.e. all constraints must be satisfied), and the constraints are somewhat independent of the original objective. A common limitation of these two methods is that the
processing order of instances influences clustering performance, and sometimes they may even fail to output a feasible partition.

To avoid this limitation, many methods treat pairwise constraints as soft constraints (i.e. a subset of these constraints could be violated) to develop more flexible approaches to embed constraints. For example, in constrained complete-link (CCL)
\cite{klein2002instance}, pairwise constraints are used to modify the instance proximity computed in the original feature space. Then, standard complete-link clustering is applied using the modified proximity matrix. Penalized probabilistic
clustering (PPC) \cite{lu2007penalized} uses pairwise constraints as a prior term w.r.t. the cluster labels within the underlying GMM-based model. Clustering configurations not satisfying the constraints have a lower probability. Moreover,
HMRF-KMEANS \cite{HMRF-constrained-2004} embeds pairwise constraints as correlations between cluster labels in a hidden Markov random field (HMRF). A metric learning step is added into standard K-means to encourage gradual satisfaction of pairwise
constraints. Other methods propagate pairwise constraints via instance similarity to obtain soft constraints, such as constrained spectral clustering \cite{lu2010constrained} and HMRF-pc \cite{wu2013constrained,wu-iccv-2013}.

\noindent\textbf{Cardinality Constraints.} They are widely used to guide the clustering process. Balancing constraints are a special type that encourages all clusters to be balanced in size or in connecting weights. For example,
normalized cuts \cite{shi2000normalized} divides the standard cut cost (sum of edge weights connecting the two clusters) by the sum of edge weights between each cluster and all other instances. Hence, each cluster is  encouraged to have similar
edge weights connecting to other clusters. Similarly, ratio cut \cite{ratio-cut-2004} normalizes the cut function by the size of each cluster to encourage similar sized clusters. Equi-sized Fuzzy c-means (FCM) \cite{klawonn2006equi} formulates the
balancing constraints as equality constraints, where the size of each cluster equals to the average cluster size. More general cardinality constraints have also been explored. For example, a constrained K-means method \cite{bradley2000constrained}
sets a lower bound on the cluster size, to avoid very small or empty clusters that occur in standard K-means. An extension of the Equi-sized FCM is proposed in \cite{hoppner2008clustering}, where the size of one single cluster is set to a specific
size.

To the best of our knowledge, the only clustering frameworks that enable both sets of constraints (Cardinality and Pairwise) either target a very specific class of methods that suffer from the locality property \cite{ding2015unified}, or are greedy heuristics that propagate constraints \cite{duong2013declarative}. Clustering methods that suffer from the locality property result in clusters located partially or entirely outside the Voronoi cell of their centers \cite{ding2015unified}. 
However, popular methods like K-means, K-medians, and many others always satisfy the locality property by definition, thus, limiting the theoretical results of \cite{ding2015unified} to a smaller class of clustering methods. There has also been an attempt to use standard constraint propagation methods to enforce both classes of constraints \cite{duong2013declarative}. However, this is done in a greedy heuristic fashion that may often fail in finding a feasible solution. Therefore, we believe that the combination of both pairwise and cardinality categories into a constraint generic and unified clustering model that can be systematically solved (ie using a flexible continuous optimization framework) has not been explored in any existing work.

\section{Proposed Method}
Unlike previous methods that can only handle either pairwise or cardinality constraints, we show, in this section, a detailed derivation of our framework that embeds both constraints simultaneously. In fact, this formulation is flexible and generic enough to handle any other linear equality. In our framework, we adopt the K-means integer program formulation \cite{peng2007approximating} expressed as follows:
\begin{equation}
\label{first_eq}
\begin{aligned}
& \min_{\{x_{ij}\}_{i=1,j=1}^{n,k}} ~~~\sum_{j=1}^{k} \sum_{i=1}^{n} x_{ij} \Big|\Big| \mathbf{s}_i - \frac{\sum_{p=1}^{n} x_{pj} \mathbf{s}_p}{\sum_{l=1}^{n} x_{lj}} \Big|\Big|_2^2 \\
& \text{s.t.} \ \sum_{j=1}^{k} x_{ij} = 1, \ \ \forall \ \ i  \ \ x_{ij} \in \{0,1 \} \ \forall \ i,j
\end{aligned}
\end{equation}
where $\mathbf{s}_p \in \mathbb{R}^{d}$ is the $p^{\text{th}}$ data point to be clustered and $k$ is the number of clusters. The variable $x_{ij}$ defines the binary association between data point $i$ and cluster $j$. The constraint $\sum_{j=1}^{k} x_{ij} = 1$ enforces data point $i$ to belong to one and only one cluster. This constraint can be simply written as a matrix vector multiplication: $\Psi^{\top} \mathbf{x} = \mathbf{1}_n$, where $\Psi^{\top} \in \mathbb{R}^{n \times nk}$ is a binary matrix that has in each row a vector $\mathbf{1}_k^\top$ that sums all the binary labels for a given data point while the rest are 0.

To simplify the fractional objective, we introduce variable $w_{pj}$, such that: $x_{pj} = w_{pj}\sum_{l=1}^{n}x_{lj}$. For ease of notation, we concatenate all the binary labels $x_{ij}$ into one vector ordered by the data points one at a time as follows: $\mathbf{x}^{\top} = \begin{bmatrix} (x_{11} & \hdots & x_{1k}) & \hdots & (x_{n1} & \hdots & x_{nk})\end{bmatrix} ^{\top}$. We also concatenate and reorder the $w_{pj}$ values one cluster at a time: $\mathbf{w}^{\top} = \begin{bmatrix} (w_{11} & \hdots & w_{n1}) & \hdots & (w_{1k} & \hdots & w_{nk})\end{bmatrix} ^{\top}$. A matrix $\mathbf{P} \in \mathbb{R}^{nk \times nk}$ is used to swap the order of the binary vectors from a cluster based order to a data point order and vice versa. Note that $\mathbf{P}$ is a proper permutation matrix that is symmetric and it satisfies: $\mathbf{P}\mathbf{P}^\top  = \mathbf{I}_{nk}$. Thus, the compact form of unconstrained K-means can be re-written as follows:

\begin{equation}
\label{second_formulation}
\begin{aligned}
& \min_{\{x_{ij}\}_{i=1,j=1}^{n,k},\{w_{pj}\}_{p=1,j=1}^{n,k}} ~~~\sum_{j=1}^{k} \sum_{i=1}^{n} x_{ij} \Big|\Big| \mathbf{s_i} - {\sum_{p=1}^{n} w_{pj} \mathbf{s_p}} \Big|\Big|_2^2  \\
& \text{s.t.} ~~\ \Psi^\top \mathbf{x} = \mathbf{1}_n, \ \ \mathbf{x} = \mathbf{P} \mathbf{w} \odot  \mathbf{C} \mathbf{x},  \ \ \mathbf{x} \in \{0,1 \}^{nk}
\end{aligned}
\end{equation}

\noindent where $\mathbf{C} \in \mathbb{R}^{nk \times nk}$ sums the binary labels of each cluster and is defined as follows:
\begin{equation}
\begin{aligned}
& \mathbf{C}^\top = \begin{bmatrix}
                 \gamma_1 &
                 \gamma_2 &
                 \dots  &
                 \gamma_n
               \end{bmatrix} \\
& \gamma_1^\top  = \begin{bmatrix}
                 1 & \mathbf{0}^\top_{k-1} & 1  & \hdots &
               \end{bmatrix}, ~\gamma_2^\top  = \begin{bmatrix}
                 0 & 1 & \mathbf{0}^\top_{k-2} & \hdots &
               \end{bmatrix} \\
\nonumber
\end{aligned}
\end{equation}

\noindent\textbf{Cardinality Constraints.} They are enforced by a set of linear constraints that specify the cluster size as:

\begin{equation}
\begin{aligned}
\label{cardinality_cons}
\sum_{i=1}^n x_{ij} = u_j~~\forall j \Leftrightarrow  \mathbf{Q} \mathbf{P}^\top \mathbf{x} = \mathbf{u} 
\end{aligned}
\end{equation}

where $u_j$ is the size of cluster $j$ and $\mathbf{Q} \in \mathbb{R}^{k \times nk}$ sums the binary labels of each cluster for all data points.

\noindent\textbf{Must-Link Constraints.} We define $\mathbf{E}_1, \mathbf{E}_2 \in \mathbb{R}^{kv \times nk}$ as selection matrices that choose the two sets of data points ($\mathbf{E}_1\mathbf{x}$ and $\mathbf{E}_2\mathbf{x}$) involved in the $v$ must-link constraints. We show next that the set of all must-link constraints  can be expressed with a single quadratic.

\begin{equation}
\begin{aligned}
\label{mustlink_cons}
\mathbf{x}^\top \mathbf{E}_1^\top \mathbf{E}_2 \mathbf{x} = v  
\end{aligned}
\end{equation}

\begin{customproposition}{1}
For the binary association $\mathbf{x} \in \{0,1\}^{nk}$ between n data points and k clusters, where $\Psi^\top \mathbf{x} = \mathbf{1}_n$, enforcing must-link constraints through $\mathbf{E}_1 \mathbf{x} = \mathbf{E}_2 \mathbf{x}$ is equivalent to enforcing a single quadratic $\mathbf{x}^\top \mathbf{E}_1^\top \mathbf{E}_2 \mathbf{x} = v$. 
\end{customproposition}

\begin{align*}
\text{\emph{Proof.}} ~ ~ ~ \mathbf{E}_1 \mathbf{x} = \mathbf{E}_2 \mathbf{x} &\Leftrightarrow \|\mathbf{E}_1\mathbf{x} - \mathbf{E}_2 \mathbf{x}\|_2^2 = 0  \\ & = \|\mathbf{E}_1 \mathbf{x}\|_2^2 + \|\mathbf{E}_2 \mathbf{x}\|_2^2 - 2 \mathbf{x}^\top \mathbf{E}_1^\top \mathbf{E}_2 \mathbf{x} = 0 \\
& = 2v - 2 \mathbf{x}^\top \mathbf{E}_1^\top \mathbf{E}_2 \mathbf{x} = 0  ~~~~~~~~~~~~~~~~~~~~~~~~ \qedsymbol
\end{align*} 
The last equality ($\|\mathbf{E}_1 \mathbf{x} \|^2 = \|\mathbf{E}_2 \mathbf{x} \|^2$ = $v$) is true since $\mathbf{x}$ is binary and that each data point is associated to only one cluster (i.e. $\Psi^\top \mathbf{x} = \mathbf{1}_n$).
\noindent This concludes that only one quadratic constraint can be used to enforce all must-link constraints.

\noindent\textbf{Cannot-Link Constraints.} We define $\mathbf{E}_3, \mathbf{E}_4 \in \mathbb{R}^{ke \times nk}$ as selection matrices for the two sets of data points ($\mathbf{E}_3\mathbf{x}$ and $\mathbf{E}_4\mathbf{x}$) involved in the $e$ cannot-link constraints. Similar to before, we show that the set of cannot-link constraints can be expressed with a single quadratic.

\begin{equation}
\begin{aligned}
\label{cannotlink_cons}
& \mathbf{x}^\top \mathbf{E}_3^\top \mathbf{E}_4 \mathbf{x} = 0 
\end{aligned}
\end{equation}

\begin{customproposition}{2}
For the binary association $\mathbf{x} \in \{0,1\}^{nk}$ between n data points and k clusters, where $\Psi^\top \mathbf{x} = \mathbf{1}_n$, enforcing cannot-link constraints through the selection matrices $\mathbf{E}_3$, $\mathbf{E}_4$ is equivalent to enforcing the quadratic $\mathbf{x}^\top \mathbf{E}_3^\top \mathbf{E}_4 \mathbf{x} = 0$. 
\end{customproposition}

\begin{align*}
\vspace{-0.35cm}
\text{\emph{Proof.}} ~ ~ ~ \text{Similar to Proposition 1 with } \|\mathbf{E}_3 \mathbf{x} + \mathbf{E}_4 \mathbf{x} \|_2^2 = 2e  ~~~~~~ \qedsymbol
\end{align*}

\noindent Incorporating Eqs (\ref{cardinality_cons}), (\ref{mustlink_cons}) and (\ref{cannotlink_cons}) into (\ref{second_formulation}) we obtain the following constrained K-means formulation:
\begin{equation}
\label{ours_admm}
\begin{aligned}
& \min_{\{x_{ij}\}_{i=1,j=1}^{n,k},\{w_{pj}\}_{p=1,j=1}^{n,k}} ~~~\sum_{j=1}^{k} \sum_{i=1}^{n} x_{ij} \Big|\Big| \mathbf{s_i} - \mathbf{S} \Lambda_j \mathbf{w}\Big|\Big|_2^2 \\
& \text{s.t.} \ \Psi^\top \mathbf{x} = \mathbf{1}_n,  \ \  \mathbf{x} \in \{0,1 \}^{nk}, \ \ \ \mathbf{Q}\mathbf{P}^\top \mathbf{x} = \mathbf{u}\\
& \mathbf{x}  =  \mathbf{P} \mathbf{w} \odot \mathbf{C} \mathbf{x} , \ \ (\mathbf{E}_1 \mathbf{x})^\top \mathbf{E}_2 \mathbf{x} = v, \ \ (\mathbf{E}_3 \mathbf{x})^\top \mathbf{E}_4 \mathbf{x} = 0
\end{aligned}
\end{equation}
\noindent where $\mathbf{S} \in \mathbb{R}^{d \times n}$ contains all the data points in its columns and $\Lambda_j \in \mathbb{R}^{n\times nk}$ is zero everywhere except for the $j^\text{th}$ block that is identity, i.e. $\Lambda_j = \begin{bmatrix} \mathbf{0} ~ \cdots & \mathbf{I}_j & \cdots ~\mathbf{0}\end{bmatrix}$.

\noindent\textbf{ADMM Solver.} 
Problem (\ref{ours_admm}) is still difficult to solve due to the mixed binary and quadratic constraints. To handle these difficulties, (i) we first replace the binary constraints with an exact equivalent set that is the intersection of the $\ell_2$-sphere (defined by set $S_2$) and box constraints (defined by set $S_b$) following \cite{Wu2016ell_pBoxAA}. (ii) Moreover, by introducing the auxiliary variables ($\mathbf{z}_1$, $\mathbf{z}_2$, $\mathbf{z}_3$, and $\mathbf{z}_4$), the quadratic constraints are now changed to bi-linear ones and separated from the binary constraints. Thus, the resultant problem is given as follows:

\begin{equation}
\label{our_all_problem}
\begin{aligned}
& \min_{\mathbf{x},\mathbf{w},\mathbf{z}_1,\mathbf{z}_2,\mathbf{z}_3,\mathbf{z}_4} ~~~~\sum_{j=1}^{k} \sum_{i=1}^{n} x_{ij} \Big|\Big| \mathbf{s_i} - \mathbf{S} \Lambda_j \mathbf{w}\Big|\Big|_2^2 \\
& \text{s.t.} \ \Psi^\top \mathbf{x} = \mathbf{1}_n,  \ \ \mathbf{z}_1 \in S_{b} , \ \ \mathbf{z}_2 \in S_{2}, \ \ \mathbf{Q}\mathbf{P}^\top \mathbf{x} = \mathbf{u}\\
& \mathbf{x}  =  \mathbf{P} \mathbf{w} \odot \mathbf{C} \mathbf{x}, \ \ (\mathbf{E}_1 \mathbf{z}_3)^\top \mathbf{E}_2 \mathbf{x} = v,\ \  (\mathbf{E}_3 \mathbf{z}_4)^\top \mathbf{E}_4 \mathbf{x} = 0 \\
& \mathbf{x} = \mathbf{z}_1 , \mathbf{x} = \mathbf{z}_2, \mathbf{x} = \mathbf{z}_3, \mathbf{x} = \mathbf{z}_4 
\end{aligned}
\end{equation}

\noindent which can be solved using in the standard ADMM framework.
Let $\mathcal{L}_{\rho_{1-9}}$
be the augmented Lagrangian function of problem (\ref{our_all_problem}). We define it as follows:

\begin{equation}
\begin{aligned}
&\mathcal{L}_{\rho_{1-9}}(\mathbf{x},\mathbf{w},\mathbf{z}_1,\mathbf{z}_2,\mathbf{z}_3,\mathbf{z}_4,\mathbf{y}_1,\mathbf{y}_2,\mathbf{y}_3,\mathbf{y}_{4},\mathbf{y}_5,y_6,\mathbf{y}_7,y_8,\mathbf{y}_9) := \\
&\sum_{j=1}^{k} \sum_{i=1}^{n} x_{ij} \Big|\Big| \mathbf{s_i} - \mathbf{S} \Lambda_j \mathbf{w}\Big|\Big|_2^2 + \mathbf{y}_1^\top (\Psi^\top \mathbf{x} - \mathbf{1}_n) + \frac{\rho_1}{2}||\Psi^\top \mathbf{x} - \mathbf{1}_n||_2^2 +  \\
&\mathbb{I}_{\{\mathbf{z}_1\in S_b\}} + \mathbf{y}_2^\top(\mathbf{x} - \mathbf{z}_1) + \frac{\rho_2}{2}||\mathbf{x} - \mathbf{z}_1||_2^2 +   \mathbb{I}_{\{\mathbf{z}_2\in S_2\}}  +  \mathbf{y}_3^\top(\mathbf{x} - \mathbf{z}_2) + \\
&\frac{\rho_3}{2}||\mathbf{x} - \mathbf{z}_2||_2^2  + \mathbf{y}_{4}^\top(\mathbf{Q} \mathbf{P}^\top \mathbf{x} - \mathbf{u}) + \frac{\rho_{4}}{2} ||\mathbf{Q}\mathbf{P}^\top \mathbf{x}- \mathbf{u} ||_2^2 + \\
& \mathbf{y}_5^\top \Big(\mathbf{I} - \text{diag}(\mathbf{P} \mathbf{w}) \mathbf{C} \Big)\mathbf{x} + \frac{\rho_5}{2}||\Big(\mathbf{I} - \text{diag}(\mathbf{P} \mathbf{w}) \mathbf{C} \Big)\mathbf{x}||_2^2 +  \\
&y_6 (\mathbf{z}_3^\top \mathbf{E}_1^\top \mathbf{E}_2 \mathbf{x} - v) +  \frac{\rho_6}{2} ||\mathbf{z}_3^\top \mathbf{E}_1^\top \mathbf{E}_2 \mathbf{x} - v ||_2^2 +  \mathbf{y}_7^\top (\mathbf{x} - \mathbf{z}_3) + \\
&\frac{\rho_7}{2} ||\mathbf{x} - \mathbf{z}_3 ||_2^2 +  y_8 (\mathbf{z}_4^\top \mathbf{E}_3^\top \mathbf{E}_4 \mathbf{x} ) +  \frac{\rho_8}{2} ||\mathbf{z}_4^\top \mathbf{E}_3^\top \mathbf{E}_4 \mathbf{x} ||_2^2 + \\
& \mathbf{y}_9^\top (\mathbf{x} - \mathbf{z}_4) + \frac{\rho_9}{2} ||\mathbf{x} - \mathbf{z}_4 ||_2^2 
\end{aligned}
\end{equation}

\noindent where the $\mathbf{y}$ variables are the Lagrange multipliers of the corresponding constraints, $\mathbb{I}$ is the indicator function that penalizes infeasible $\mathbf{z}_1$ and $\mathbf{z}_2$, and $\rho_{1-9} \ge 0$ are the penalty parameters. In our experiments, we set all the $\rho$ coefficients to the same value. The iterative ADMM steps for problem (\ref{our_all_problem}) are described in Algorithm \ref{algo}. ADMM updates are performed by optimizing for the set of primal variables one at a time, while keeping the rest of the primal and dual variables fixed. Then, the dual variables are updated using gradient ascent on the corresponding dual problem.

\noindent We next show the final updates for each subproblem, but the exact derivations are found in the \textbf{supplementary material}.
\begin{algorithm}[t]
\label{algo}
\caption{ADMM for Solving Problem (\ref{our_all_problem})}
\SetKwInOut{Input}{Input}
\SetKwInOut{Output}{Output}
\Input{Set $\mathbf{S} \in \mathbb{R}^{d \times n}$. Set $\rho_{1-9}$, $\mathbf{y}_{1-5,7,9} = \mathbf{0}$, $y_{6,8} = 0$, $\mathbf{x}_{\text{kmeans}}$, $\mathbf{w} = \mathbf{P}^\top\mathbf{x} \odot \text{diag}^{-1}(\mathbf{C} \mathbf{x}) \mathbf{1}_{nk}$.}
\Output{$\mathbf{x}$}
  \While{\emph{not converged}}{
 \textbf{update:} $\mathbf{x}$ by solving Eq (\ref{eq:update_x}). \newline
 \textbf{update:} $\mathbf{w}$ by solving Eq (\ref{eq:update_w}). \newline
 \textbf{update:} $\mathbf{z}_{1-4}$ via Eqs (\ref{update_z1},\ref{update_z2}, \ref{update_z3},\ref{update_z4}). \newline
 \textbf{update:} $\mathbf{y}_{1-5,7,9},y_{6,8}$ via Eqs (\ref{eq:all_ys}).
  }
\end{algorithm}

\paragraph{Update $\mathbf{x}$ :} We need to solve the following linear system using the conjugate gradient method.

\begin{equation}
\begin{aligned}
\label{eq:update_x}
&\Big( \rho_1 \Psi \Psi^\top + (\rho_2 + \rho_3 + \rho_7 + \rho_9) \mathbf{I}_{nk} + \rho_{4} \mathbf{P} \mathbf{Q}^\top \mathbf{Q} \mathbf{P}^\top  + \\
&\rho_5 \Big(\mathbf{I} - \text{diag}(\mathbf{P}\mathbf{w})\mathbf{C} \Big)^\top \Big(\mathbf{I} - \text{diag}(\mathbf{P}\mathbf{w})\mathbf{C}\Big) + \\
& \rho_6 \mathbf{E}_2^\top \mathbf{E}_1 \mathbf{z}_3 \mathbf{z}_3^\top \mathbf{E}_1^\top \mathbf{E}_2 + \rho_8 \mathbf{E}_4^\top \mathbf{E}_4 \mathbf{z}_4 \mathbf{z}_5^\top \mathbf{E}_3^\top \mathbf{E}_4 \Big) \mathbf{x} = \\
& - \Big( \text{vect}(\mathbf{B}) +   \Psi  \mathbf{y}_1 + \mathbf{y}_2 + \mathbf{y}_3  - \rho_1 \Psi \mathbf{1}_{n} - \rho_2 \mathbf{z}_1 - \\
& \rho_3 \mathbf{z}_2 - \mathbf{C}^\top \text{diag}(\mathbf{P} \mathbf{w}) \mathbf{y}_5 + y_6 \mathbf{E}_2^\top \mathbf{E}_1 \mathbf{z}_4 -  \rho_6 v \mathbf{E}_2^\top \mathbf{E}_1 \mathbf{z}_3 + \mathbf{y}_7 - \\
&\rho_7 \mathbf{z}_3 + y_8 \mathbf{E}_4^\top \mathbf{E}_3 \mathbf{z}_4 + \mathbf{y}_9 -\rho_9 \mathbf{z}_4 + \mathbf{P} \mathbf{Q}^\top \mathbf{y}_{4} - \rho_{4} \mathbf{P} \mathbf{Q}^\top \mathbf{u}  \Big) 
\end{aligned}
\end{equation}
where $\mathbf{B}(i,j) = \|\mathbf{s_i} - \mathbf{S} \Lambda_j \mathbf{w} \|_2^2$, and $\text{vect}(\mathbf{B})$ is simply a columnwise vectorization of the matrix $\mathbf{B}$.

\paragraph{Update $\mathbf{w}$:} We need to solve the following linear system using the conjugate gradient method.

\begin{equation}
\begin{aligned}
\label{eq:update_w}
&\Big [ \sum_j^k \sum_i^n 2x_{ij} \Lambda_j^\top \mathbf{S}^\top \mathbf{S} \Lambda_j   + \rho_5 \mathbf{P}^\top  \text{diag} \Big( \mathbf{C} \mathbf{x} \odot \mathbf{C}\mathbf{x} \Big) \mathbf{P}  \Big] \mathbf{w} \\
& = \sum_j^k \sum_i^n 2 x_{ij} \Lambda_j^\top \mathbf{S}^\top \mathbf{s}_i  + \mathbf{P}^\top \mathbf{C} \mathbf{x} \odot \mathbf{y}_5 + \rho_5 \mathbf{P}^\top \mathbf{C} \mathbf{x} \odot \mathbf{x} 
\end{aligned}
\end{equation}

\paragraph{Update $\mathbf{z}_1$:} Here, we need to perform a simple projection onto the box: $S_b=\{\mathbf{a}: \mathbf{0}\leq\mathbf{a}\leq\mathbf{1}\}$. This projection is an elementwise clamping between $0$ and $+1$.

\begin{equation}
\label{update_z1}
\mathbf{z}_1 = \mathbf{P}_{S_b}\left(\mathbf{x}+\frac{\mathbf{y}_2}{\rho_2}\right)=\min\left(\max\left(\mathbf{x}+\frac{\mathbf{y}_2}{\rho_2},\mathbf{0}\right),\mathbf{1}\right)
\end{equation}

\paragraph{Update $\mathbf{z}_2$:} We need to perform a simple projection onto the $\ell_2$-sphere: $S_2=\{\mathbf{a}\in\mathbb{R}^{nk}: \|\mathbf{a}-\frac{1}{2}\mathbf{1}\|_2^2=\frac{nk}{4}\}$. This involves an elementwise shift and $\ell_2$ vector normalization.

\begin{equation}
\label{update_z2}
\mathbf{z}_2 = \mathbf{P}_{S_2}\left(\mathbf{x}+\frac{\mathbf{y}_3}{\rho_3}\right)=\frac{\sqrt{nk}}{2}\frac{\left(\mathbf{x}+\frac{\mathbf{y}_3}{\rho_3}\right)-\frac{1}{2}\mathbf{1}}{\left\|\left(\mathbf{x}+\frac{\mathbf{y}_3}{\rho_3}\right)-\frac{1}{2}\mathbf{1}\right\|_2}+\frac{1}{2}\mathbf{1}
\end{equation}

\paragraph{Update $\mathbf{z}_3$:}  We need to solve the following linear system using the conjugate gradient method.

\begin{align}
&\Big [\rho_6 \mathbf{E}_1^\top \mathbf{E}_2 \mathbf{x} \mathbf{x}^\top \mathbf{E}_2^\top \mathbf{E}_1  + \rho_7 \mathbf{I}_{nk}\Big]\mathbf{z}_3 =  \mathbf{y}_7 + \rho_7 \mathbf{x} - y_6 \mathbf{E}_1^\top \mathbf{E}_2 \mathbf{x} + \notag \\
& \ \ \ \ \ \ \ \ \ \ \ \rho_6 v  \mathbf{E}_1^\top \mathbf{E}_2 \mathbf{x} 
\label{update_z3} 
\end{align}

\paragraph{Update $\mathbf{z}_4$:} We need to solve the following linear system using the conjugate gradient method.

\begin{equation}
\label{update_z4}
\Big [ \rho_8 \mathbf{E}_3^\top \mathbf{E}_4 \mathbf{x} \mathbf{x}^\top \mathbf{E}_4^\top \mathbf{E}_3 + \rho_9 \mathbf{I}_{nk}\Big] \mathbf{z}_4 = \mathbf{y}_9 + \rho_9 \mathbf{x} - y_8 \mathbf{E}_3^\top \mathbf{E}_4 \mathbf{x} \end{equation}

\paragraph{Update $\mathbf{y}_1,\mathbf{y}_2,\mathbf{y}_3,\mathbf{y}_{4},\mathbf{y}_5,y_6,\mathbf{y}_7,y_8,\mathbf{y}_9$:} Lastly, we need to perform dual ascent on the dual variables as follows:

\begin{equation}
\begin{aligned}
\label{eq:all_ys}
& \mathbf{y}_1 \leftarrow \mathbf{y}_1 + \rho_1 \Big(  \Psi^\top \mathbf{x} - \mathbf{1}_n \Big), \ \ \ \mathbf{y}_2 \leftarrow \mathbf{y}_2 + \rho_2 \Big( \mathbf{x} - \mathbf{z}_2 \Big) \\
& \mathbf{y}_3 \leftarrow \mathbf{y}_3 + \rho_3 \Big( \mathbf{x} - \mathbf{z}_3 \Big), \ \ \ \mathbf{y}_4 \leftarrow \mathbf{y}_{4} + \rho_{4} \Big( \mathbf{Q} \mathbf{P}^\top \mathbf{x} - \mathbf{u} \Big) \\
& \mathbf{y}_5 \leftarrow \mathbf{y}_5 + \rho_5 \Big( \mathbf{x} - \mathbf{P} \mathbf{w} \odot \mathbf{C} \mathbf{x} \Big), \ \ \ y_6 \leftarrow y_6 + \rho_6 \Big( \mathbf{z}_3^\top \mathbf{E}_1^\top \mathbf{E}_2 \mathbf{x} - v \Big) \\
& \mathbf{y}_7 \leftarrow \mathbf{y}_7 + \rho_7 \Big( \mathbf{x} - \mathbf{z}_3 \Big), \ \ \  y_8 \leftarrow y_8 + \rho_8 \Big( \mathbf{z}_4^\top \mathbf{E}_3^\top \mathbf{E}_4 \mathbf{x} \Big) \\
& \mathbf{y}_9 \leftarrow \mathbf{y}_9 + \rho_9 \Big( \mathbf{x} - \mathbf{z}_4 \Big) 
\end{aligned}
\end{equation}

\noindent The ADMM iterations are run until convergence (i.e. when the standard deviation between the last 10 objective values is $\leq {10}^{-5}$). Upon convergence, all the primal variables ($\mathbf{x}$ and $\mathbf{z}_{1-4}$) converge to the same feasible binary vector. Despite that the problem is non-convex, we show empirically in the experiments' section and in the \textbf{supplementary material} that the performance using is very stable.

\section{Experiments}
In this section, we conduct extensive experiments to motivate and evaluate our proposed clustering method, both on synthetic and real datasets. We also compare our method against other constrained clustering methods on well-known benchmarks, thus,
demonstrating superior performance and flexibility, as well as, superior gain that can be achieved when both categories of constraints (cardinality and pairwise) are combined in our framework.

\begin{figure}[t]
\begin{center}
\scalebox{0.70}{
\includegraphics[width=0.48\textwidth]{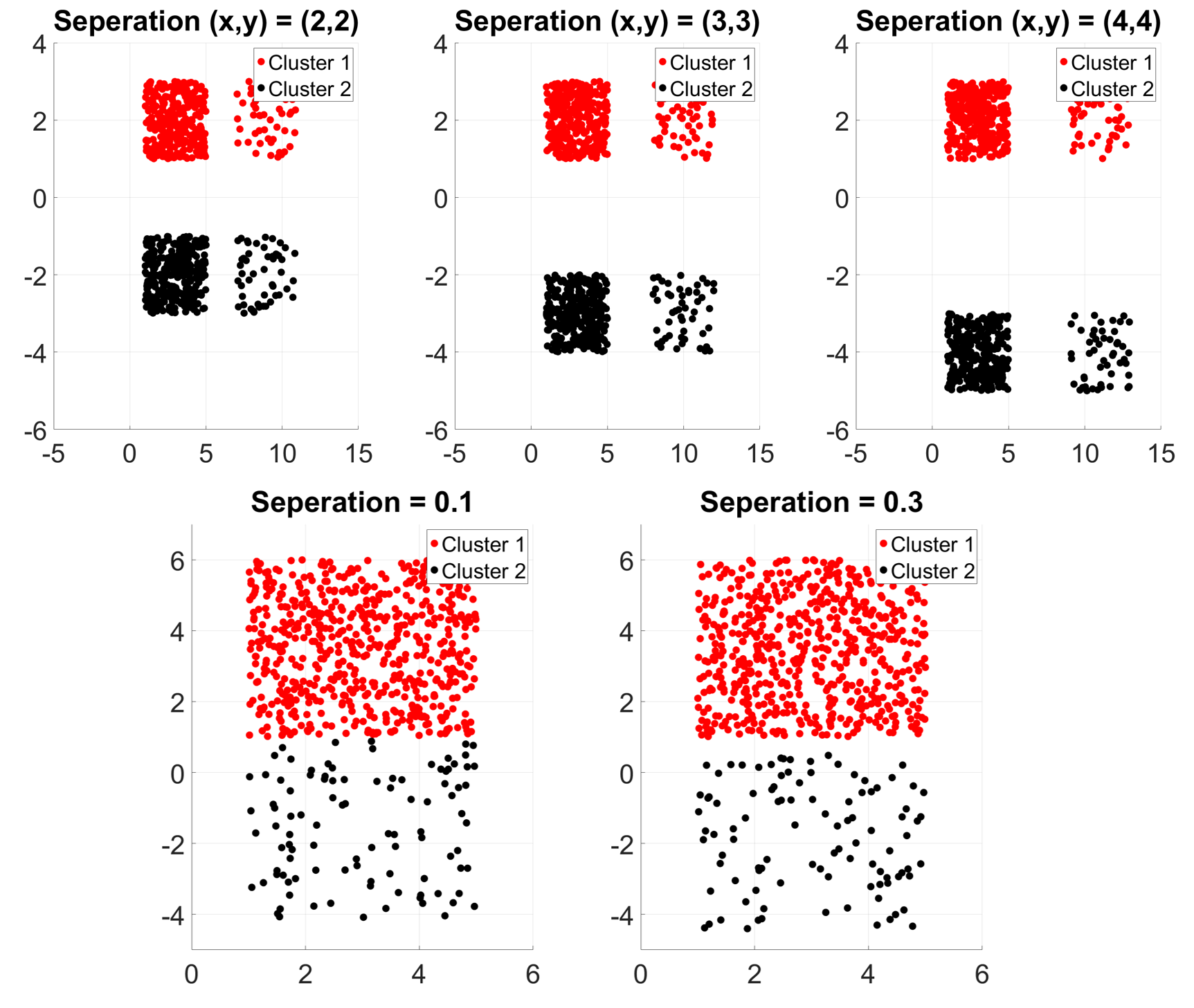}
}
\end{center}
\caption{The \texttt{\emph{Balanced}} and  \texttt{\emph{ImBalanced}} in the two consecutive rows respectively. They comprise two clusters (red/black) with an increasing separation
between clusters.}\label{fig:balanced_imbalanced}
\end{figure}

\noindent\textbf{1. Datasets and Implementation Details.} The datasets used in this section vary from synthetic to real. As for the synthetic ones, we construct two datasets, one is cluster balanced (denoted as \texttt{\emph{Balanced}})
and the other is imbalanced (denoted as \texttt{\emph{ImBalanced}}) as shown in Figure \ref{fig:balanced_imbalanced}. Each dataset comprises 700 data points with 2 clusters. In \texttt{\emph{Balanced}}, each cluster has exactly 350 data points,
while in \texttt{\emph{ImBalanced}} one cluster has 600 data points while the other contains 100. As for the real datasets, we make use of various popular UCI datasets \cite{bradley2000constrained}, e.g. iris, wine, glass, ionosphere, Hepatitis,
Hepatitis1 and Breast Cancer Wis-D. These are the most popular UCI datasets used for clustering purposes \cite{wagstaff2001constrained,lu2010constrained}. Following convention, data points are normalized to have a value in $[-1,+1]$. For Hepatitis
and Hepatitis1, we remove all points with missing or none categorical features. Table \ref{datasets} lists the details of all UCI datasets used in the experiments.

\begin{table}[t]
\scriptsize
\centering
\caption{Lists the total number of points, clusters and features of all UCI datasets used in the experiments.}
\label{datasets}
\begin{tabular}{|c|c|c|c|}
\hline
\toprule
Datasets                             & $\#$Points       & $\#$Features       & $\#$Clusters  \\ \hline
\midrule
wine                                 & 178             & 13                & 3   \\ \hline
iris                                 & 150             & 4                 & 3   \\ \hline
glass                                & 214             & 9                 & 7   \\ \hline
ionosphere                           & 351             & 33                & 2   \\ \hline
Hepatitis                            & 142             & 14                & 2   \\ \hline
Hepatitis1                           & 80              & 19                & 2  \\ \hline
Breast Cancer Wis-D                  & 569             & 30                & 2  \\ \hline
\bottomrule
\end{tabular}
\end{table}

\begin{figure}[!htp]
\begin{center}
\includegraphics[width=0.5\textwidth]{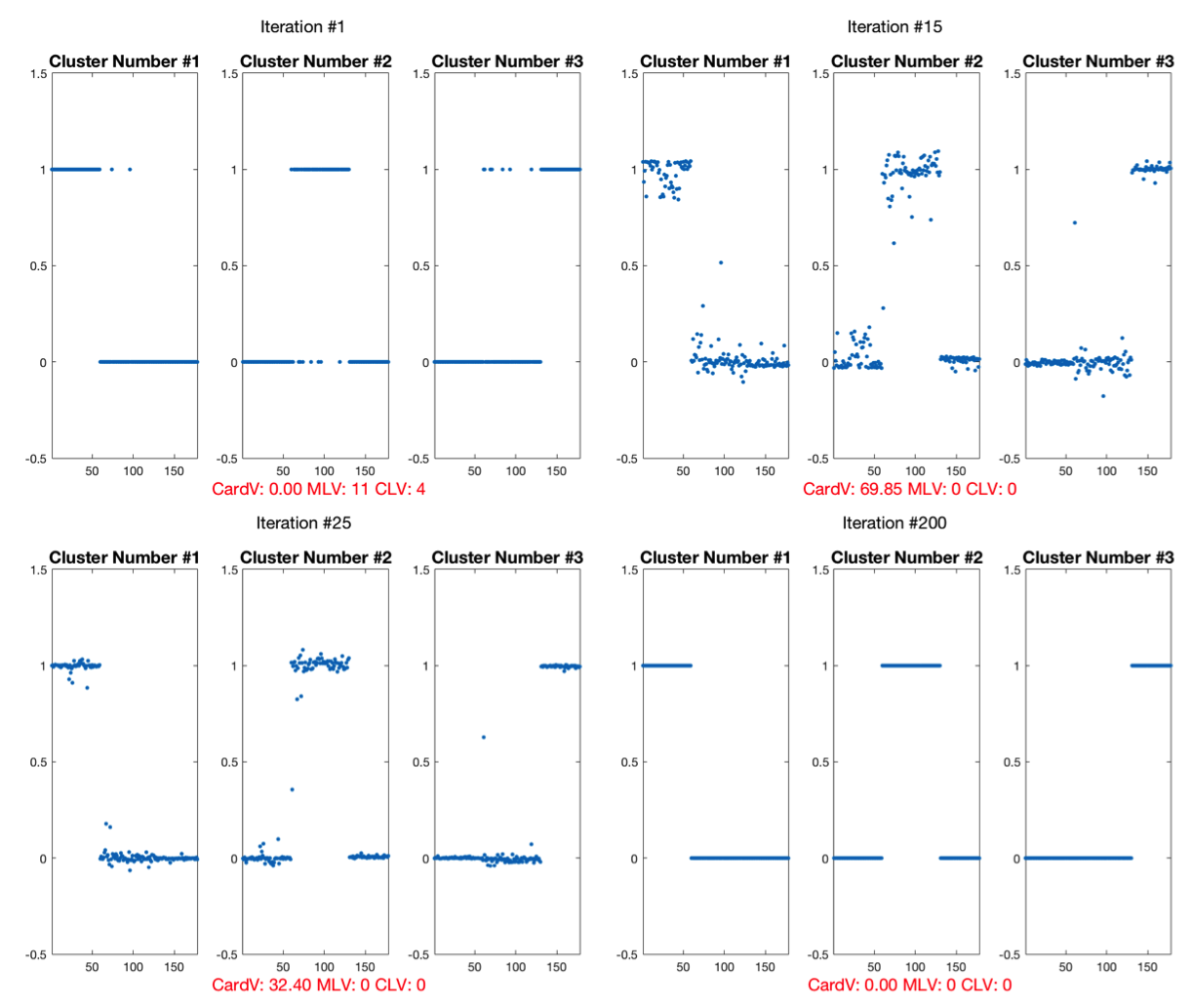}
\caption{Convergence of the solution $\mathbf{x}$ using $\ell_p$Km-Mix with random binary initialization that satisfy the cardinality constraints on the Wine dataset.
}\label{fig: convergence constrained x}
\end{center}
\end{figure}

\begin{table}[t]
\scriptsize
\centering
\caption{Comparison between K-means and $\ell_p$Km on real UCI datasets using K-means
objective value, \texttt{\emph{ARI}}(\%), \texttt{\emph{MI}} and \texttt{\emph{HI}}(\%) along with the standard deviation.}
\label{kmeans_oursuncon}
\begin{tabular}{|c|c|c|}
\hline
\toprule
Datasets        & K-mean (Obj Value)               & $\ell_p$Km (Obj Value)  \\ \hline
\midrule
wine            &  195.91  $\pm$ 0.05              & \textbf{195.86} $\pm$ 0.07   \\ \hline
iris            &  29.894  $\pm$ 4.84              & \textbf{28.282} $\pm$ 0.05    \\ \hline
glass           &  \textbf{81.277}  $\pm$ 5.49     & 85.944 $\pm$ 1.307        \\ \hline
\midrule
                & K-means (\texttt{\emph{ARI}})   & $\ell_p$Km (\texttt{\emph{ARI}}(\%))  \\ \hline
wine            &  84.765 $\pm$ 0.92              &  \textbf{86.779} $\pm$ 0.10     \\ \hline
iris            &  67.831 $\pm$ 8.79              &  \textbf{72.41} $\pm$  0.90     \\ \hline
glass           &  \textbf{17.454} $\pm$ 1.31     & 14.96 $\pm$  0.91       \\ \hline
\midrule
                & K-means (\texttt{\emph{MI}})        & $\ell_p$Km (\texttt{\emph{MI}})  \\ \hline
wine            &  0.068  $\pm$ 0.004               &  \textbf{0.059} $\pm$ 0.001     \\ \hline
iris            &  0.145  $\pm$ 0.048               &  \textbf{0.122} $\pm$ 0.004     \\ \hline
glass           &  0.338 $\pm$ 0.019                &  \textbf{0.319} $\pm$ 0.019       \\ \hline
\midrule
                & K-means (\texttt{\emph{HI}})        & $\ell_p$Km (\texttt{\emph{HI}}(\%))  \\ \hline
wine            &  86.423  $\pm$ 0.82                 &  \textbf{88.225} $\pm$ 0.001     \\ \hline
iris            &  70.969  $\pm$ 9.51                 &  \textbf{75.590} $\pm$ 0.008     \\ \hline
glass           &  32.378  $\pm$ 3.57                 &  \textbf{36.134} $\pm$ 0.038       \\ \hline
\bottomrule
\end{tabular}
\end{table}

\begin{table}[t]
\scriptsize
\centering
\caption{Comparison between K-means and $\ell_p$Km-Car on synthetic balanced and Imbalaced synthetic datasets using \texttt{\emph{ARI}}(\%), \texttt{\emph{MI}} and \texttt{\emph{HI}}(\%).}
\label{kmeans_card}
\begin{tabular}{|c|c|c|}
\hline
\toprule
Datasets                   & K-mean (\texttt{\emph{ARI}})           & $\ell_p$Km-Car (\texttt{\emph{ARI}})  \\ \hline
\midrule
Balanced (x,y)=(2,2)       &  59.97  $\pm$ 51.68                    & \textbf{100}   $\pm$ 0    \\ \hline
Balanced (x,y)=(3,3)       &  89.99  $\pm$ 31.64                    & \textbf{100}   $\pm$ 0    \\ \hline
Balanced (x,y)=(4,4)       &  79.99  $\pm$ 42.19                    & \textbf{100}   $\pm$ 0    \\ \hline
Imbalanced y=0.1           &  34.67  $\pm$ 0                        & \textbf{99.46} $\pm$ 0.37   \\ \hline
Imbalanced y=0.3           &  32.26  $\pm$ 0                        & \textbf{100}   $\pm$ 0             \\ \hline
\midrule
                                & K-means (\texttt{\emph{MI}})      & $\ell_p$Km-Car (\texttt{\emph{MI}})  \\ \hline
Balanced (x,y)=(2,2)            &  0.2003  $\pm$ 0.2586             & \textbf{0} $\pm$ 0   \\ \hline
Balanced (x,y)=(3,3)            &  0.0501  $\pm$ 0.1583             & \textbf{0} $\pm$ 0    \\ \hline
Balanced (x,y)=(4,4)            &  0.1001  $\pm$ 0.2111             & \textbf{0} $\pm$ 0    \\ \hline
Imbalanced y=0.1                &  0.3082  $\pm$ 0                  & \textbf{0.002} $\pm$ 0.0014        \\ \hline
Imbalanced y=0.3                &  0.3222  $\pm$ 0                  & \textbf{0} $\pm$ 0        \\ \hline
\midrule
                                & K-means (\texttt{\emph{HI}})      & $\ell_p$Km-Car (\texttt{\emph{HI}})  \\ \hline
Balanced (x,y)=(2,2)            &  59.94  $\pm$ 5.171               & \textbf{100} $\pm$ 0   \\ \hline
Balanced (x,y)=(3,3)            &  89.99  $\pm$ 31.67               & \textbf{100} $\pm$ 0    \\ \hline
Balanced (x,y)=(4,4)            &  79.97  $\pm$ 42.22               & \textbf{100} $\pm$ 0    \\ \hline
Imbalanced y=0.1                &  38.35  $\pm$ 0                   & \textbf{99.6}$\pm$ 0.28        \\ \hline
Imbalanced y=0.3                &  35.57  $\pm$ 0                   & \textbf{100} $\pm$ 0        \\ \hline
\bottomrule
\end{tabular}
\end{table}

As for the implementation details, none of the selection matrices used in the proposed framework (i.e. $\mathbf{E}_1,\mathbf{E}_2, \mathbf{E}_3, \mathbf{E}_4, \mathbf{P}, \mathbf{C}, \mathbf{Q}, \Psi, \Lambda_j$) are actually constructed.
Only element indexing within vectors is used, thus, keeping the necessary computation cost minimal. For ease, all $\rho_i$ parameters have the same value and updated similarly. We find that setting all $\rho_i$ parameters to 20 and by increasing it
it every 5 iterations by 10\% for all real datasets achieves the fastest convergence. Moreover, we initialize all the optimization variables using zero vectors, while $\mathbf{x}$ is initialized to random (i.e. random assignment of data points to
clusters) if the comparison is against K-means. When comparing against other clustering methods, we use the same K-means initialization as other methods. In all comparisons, $\mathbf{w}$ is initialized to a feasible point as given in Algorithm
\ref{algo}. MATLAB is used to implement our method.
The most expensive operation in our framework is the $\mathbf{x}$ and $\mathbf{w}$  updates, which involve solving an $n \times k$ linear system. This is the bottleneck of our
framework causing it to have a computational complexity  $\mathcal{O}(n^3 k^3)$ per iteration. 
In the final experiment, we report the runtime of our framework on different sized datasets with a variety of constraint choices.

As for the evaluation metric, we adopt  the 3 most common criteria used in the clustering community to compare different clustering methods, namely the Adjusted Random Index (\texttt{\emph{ARI}})($\nearrow$), Mirkin's Index (\texttt{\emph{MI}})
($\searrow$) and Hubert's Index (\texttt{\emph{HI}}) ($\nearrow$) which calculate a measure of agreement between two partitions of a dataset \cite{Hubert1985,meilua2007comparing}. The symbol $\nearrow$ indicate that the higher the number the better
performance and vice versa for $\searrow$. In all experiments, clustering is repeated 10 times with different initializations and we report the average and standard deviation of the metric used in comparison.


\noindent\textbf{2. Comparing Different Constraint Design Choices.} We apply our proposed method, $\ell_p$Km, on the same clustering task with several choices of constraints: no constraints, only cardinality constraints, only pairwise
constraints and both types jointly. We refer to each as $\ell_p$Km, $\ell_p$Km-Car, $\ell_p$Km-Pair and $\ell_p$Km-Mix respectively.

\noindent\textbf{(i) An Auxiliary Experiment.} Despite that we do not provide a proof for the convergence of the non-quadratic objective in Eq. \ref{first_eq}, as it is proven for the quadratic case in \cite{Wu2016ell_pBoxAA}, we find
the performance very stable where we demonstrate it empirically. For instance, we run $\ell_p$Km-Mix that enforces cardinality, 20 must-link and 20 cannot-link constraints. In figure \ref{fig: convergence constrained x}, we plot the three pieces of
the solution label vector $\mathbf{x}$ at four different ADMM iterations (1, 15, 25, and 200). In the first iteration, the initial clustering is random however satisfying the cardinality constraints, so it is binary but it does not lead to a good
objective. As ADMM progresses, the continuous solution $\mathbf{x}$ becomes more and more binary, until it converges to a feasible binary solution where the three clusters are disjoint satisfying all constraints. Moreover, we also report the number
of cardinality (CardV), must-link (MLV), and cannot-link (CLV) violations at each of these iterations. These violations gradually decrease until convergence occurs, when no violations persist. We find this stable performance across all datasets as
will be presented in later sections. Further detailed experiments can be found in the \textbf{supplementary material}.

\begin{table*}[!htb]
\scriptsize
\centering
\caption{Comparison of several pairwise constrained clustering methods against $\ell_p$Km-Pair using \texttt{\emph{ARI(\%)}}, as well as, must-link (MLV) and cannot-link (CLV) violations in
the constraints. The cells indicated with x imply that the underlying method does not attain a feasible solution after 1000 runs.}
\label{table:pair_com1}
\scalebox{0.90}{
\begin{tabular}{|c|c|ccc|ccc|ccc|ccc|ccc|}
\cmidrule[1.1pt]{1-17}
& {\multirow{2}{*}{Constraints}} & \multicolumn{3}{c}{COP-KMEANS} &  \multicolumn{3}{|c}{Spectral Clustering}     & \multicolumn{3}{|c}{PPC}             & \multicolumn{3}{|c}{CCL}                & \multicolumn{3}{|c|}{$\ell_p$Km-Pair}
 \\
 \\[-8pt]
  \cline{3-17}
\\[-7pt]
 &     &   \multicolumn{1}{c}{ARI($\%$)} & \multicolumn{1}{c}{MLV} & \multicolumn{1}{c}{CLV}         & \multicolumn{1}{|c}{ARI($\%$)} & \multicolumn{1}{c}{MLV} & \multicolumn{1}{c}{CLV}        & \multicolumn{1}{|c}{ARI($\%$)} & \multicolumn{1}{c}{MLV} & \multicolumn{1}{c}{CLV}                   & \multicolumn{1}{|c}{ARI($\%$)} & \multicolumn{1}{c}{MLV} & \multicolumn{1}{c}{CLV}                          & \multicolumn{1}{|c}{ARI($\%$)} & \multicolumn{1}{c}{MLV} & \multicolumn{1}{c|}{CLV} \\
\\[-8pt]
\hline
\parbox[t]{2mm}{\multirow{5}{*}{\rotatebox[origin=c]{90}{wine}}}
\\[-6pt]
 & 20ml, 20cl          & 87.14 $\pm$ 1.06          & 0 & 0             & 91.07 $\pm$ 2            & 0  & 0    & 88.19 $\pm$ 0      & 1    & 0    & 74.52  $\pm$ 0   & 0    & 0    & \textbf{91.67}  $\pm$ 0  & 0  & 0  \\
 & 40ml, 40cl          & 84.48 $\pm$ 0.71          & 0 & 0             & \textbf{96.67} $\pm$ 0   & 0  & 0    & 89.68 $\pm$ 1.64   & 1    & 1.5  & 53.58  $\pm$ 0   & 3    & 4    & 88.04           $\pm$ 0  & 0  & 0      \\
 & 60ml, 60cl          & 91.22 $\pm$ 0             & 0 & 0             & 94.87 $\pm$ 0            & 2  & 0    & 89.74 $\pm$ 0.04   & 4.8  & 2.9  & 82.79  $\pm$ 0   & 7    & 1    & \textbf{98.32}  $\pm$ 0  & 0  & 0    \\
 & 80ml, 80cl          & 93.09 $\pm$ 0             & 0 & 0             & \textbf{96.67} $\pm$ 0   & 1  & 0    & 87.73 $\pm$ 0.74   & 6    & 3    & 55     $\pm$ 0   & 11   & 10   & 94.87           $\pm$ 0  & 0  & 0    \\
 & 100ml, 100cl        & \textbf{96.51} $\pm$ 0    & 0 & 0             & 94.87 $\pm$ 0            & 3  & 2    & 88.33 $\pm$ 0.14   & 4.4  & 2.1  & 90.34  $\pm$ 0   & 1    & 5    & \textbf{96.51}  $\pm$ 0  & 0  & 0    \\
\hline
\parbox[t]{2mm}{\multirow{5}{*}{\rotatebox[origin=c]{90}{iris}}}
\\[-6pt]
 & 20ml, 20cl          & 69.57 $\pm$ 0.61                                           & 0                                           & 0                                              & 80.27 $\pm$ 0.64    & 1.1   & 2.2    & \textbf{90.39}  $\pm$ 0  & 1  & 0    & 68.28 $\pm$ 0    & 0    & 2     & 72.87  $\pm$ 0  & 0   & 0    \\
 & 40ml, 40cl          & 74.37 $\pm$ 0                                              & 0                                           & 0                                              & 90.38 $\pm$ 0       & 1     & 1      & \textbf{92.22}  $\pm$ 0  & 1  & 2    & 45.65 $\pm$ 0    & 3    & 11    & 72.76  $\pm$ 0  & 0   & 0    \\
 & 60ml, 60cl          & \textcolor[rgb]{0.88,0.13,0.21}{\textbf{x}}                & \textcolor[rgb]{0.88,0.13,0.21}{\textbf{x}} & \textcolor[rgb]{0.88,0.13,0.21}{\textbf{x}}    & 69.00 $\pm$ 0.59    & 0     & 19.5   & 97.99           $\pm$ 0  & 0  & 1    & 54.86 $\pm$ 0    & 2    & 0     & \textbf{100}    $\pm$ 0  & 0   & 0    \\
 & 80ml, 80cl          & \textcolor[rgb]{0.88,0.13,0.21}{\textbf{x}}                & \textcolor[rgb]{0.88,0.13,0.21}{\textbf{x}} & \textcolor[rgb]{0.88,0.13,0.21}{\textbf{x}}    & 68.44 $\pm$ 0       & 0     & 20     & 97.99           $\pm$ 0  & 0  & 1    & 54.17 $\pm$ 0    & 2    & 0     & \textbf{100}    $\pm$ 0  & 0   & 0    \\
 & 100ml, 100cl        & \textcolor[rgb]{0.88,0.13,0.21}{\textbf{x}}                & \textcolor[rgb]{0.88,0.13,0.21}{\textbf{x}} & \textcolor[rgb]{0.88,0.13,0.21}{\textbf{x}}    & 68.44 $\pm$ 0       & 0     & 20     & 97.99           $\pm$ 0  & 0  & 1    & 52.52 $\pm$ 0    & 4    & 1     & \textbf{100}    $\pm$ 0  & 0   & 0    \\
\cmidrule[1.1pt]{1-17}
\end{tabular}}
\end{table*}

\begin{table*}[!htb]
\scriptsize
\centering
\caption{Comparison of several pairwise constrained clustering methods against $\ell_p$Km-Pair using \texttt{\emph{MI}} and \texttt{\emph{HI(\%)}}. The cells indicated with x imply that the underlying method does not attain a feasible solution after 1000 runs.}
\label{table:pair_com2}
\scalebox{0.9}{
\begin{tabular}{|c|c|cc|cc|cc|cc|cc|}
\cmidrule[1.1pt]{1-12}
& {\multirow{2}{*}{Constraints}} & \multicolumn{2}{c}{COP-KMEANS} &  \multicolumn{2}{|c}{Spectral Clustering}     & \multicolumn{2}{|c}{PPC}             & \multicolumn{2}{|c}{CCL}                & \multicolumn{2}{|c|}{$\ell_p$Km-Pair}
 \\
 \\[-8pt]
  \cline{3-12}
\\[-7pt]
 &  & \multicolumn{1}{c}{MI} & \multicolumn{1}{c}{HI($\%$)}  & \multicolumn{1}{|c}{MI} & \multicolumn{1}{c}{HI($\%$)}  & \multicolumn{1}{|c}{MI} & \multicolumn{1}{c}{HI($\%$)}& \multicolumn{1}{|c}{MI} & \multicolumn{1}{c}{HI($\%$)}          & \multicolumn{1}{|c}{MI} & \multicolumn{1}{c|}{HI($\%$)}   \\
\\[-8pt]
\hline
\parbox[t]{2mm}{\multirow{5}{*}{\rotatebox[origin=c]{90}{wine}}}
\\[-6pt]
 & 20ml, 20cl          & 0.06 $\pm$ 0.01 & 88.55 $\pm$ 0.95           & 0.04 $\pm$ 0.01   & 92.03 $\pm$ 1.8             & 0.05 $\pm$ 0    & 89.48 $\pm$ 0       & 0.11  $\pm$ 0   & 77.32 $\pm$ 0     & 0.04  $\pm$ 0  & \textbf{92.57} $\pm$ 0     \\
 & 40ml, 40cl          & 0.07 $\pm$ 0    & 86.18 $\pm$ 0.62           & 0.02 $\pm$ 0      & \textbf{97.03} $\pm$ 0      & 0.05 $\pm$ 0.02 & 90.80 $\pm$ 1.46    & 0.21  $\pm$ 0   & 57.37 $\pm$ 0     & 0.05  $\pm$ 0  & 89.34 $\pm$ 0         \\
 & 60ml, 60cl          & 0.04 $\pm$ 0    & 92.15 $\pm$ 0              & 0.02 $\pm$ 0      & 95.42 $\pm$ 0               & 0.05 $\pm$ 0    & 90.85 $\pm$ 0.04    & 0.08  $\pm$ 0   & 84.69 $\pm$ 0     & 0.01  $\pm$ 0  & \textbf{98.50} $\pm$ 0      \\
 & 80ml, 80cl          & 0.03 $\pm$ 0    & 93.83 $\pm$ 0              & 0.02 $\pm$ 0      & \textbf{97.03} $\pm$ 0      & 0.06 $\pm$ 0    & 89.07 $\pm$ 0.66    & 0.20  $\pm$ 0   & 59.63 $\pm$ 0     & 0.02  $\pm$ 0  & 95.42 $\pm$ 0       \\
 & 100ml, 100cl        & 0.02 $\pm$ 0    & \textbf{96.88} $\pm$ 0     & 0.02 $\pm$ 0      & 95.42 $\pm$ 0               & 0.05 $\pm$ 0    & 89.60 $\pm$ 0.13    & 0.04  $\pm$ 0   & 91.39 $\pm$ 0     & 0.02  $\pm$ 0  & \textbf{96.88} $\pm$ 0       \\
\hline
\parbox[t]{2mm}{\multirow{5}{*}{\rotatebox[origin=c]{90}{iris}}}
\\[-6pt]
 & 20ml, 20cl          & 0.14 $\pm$ 0.01                                           & 72.92 $\pm$ 0.49                                   & 0.09          $\pm$ 0.036  & 82.41          $\pm$ 5.11   & \textbf{0.04}  $\pm$ 0  & \textbf{91.50} $\pm$ 0     & 0.14 $\pm$ 0    & 71.36 $\pm$ 0      & 0.12        $\pm$ 0  & 75.95 $\pm$ 0        \\
 & 40ml, 40cl          & 0.11 $\pm$ 0                                              & 77.18 $\pm$ 0.75                                   & \textbf{0.04} $\pm$ 0      & \textbf{91.50} $\pm$ 0      & 0.03           $\pm$ 0  & 93.13 $\pm$ 0              & 0.26 $\pm$ 0    & 49.01 $\pm$ 0      & 0.12        $\pm$ 0  & 75.95 $\pm$ 0        \\
 & 60ml, 60cl          & \textcolor[rgb]{0.88,0.13,0.21}{\textbf{x}}               & \textcolor[rgb]{0.88,0.13,0.21}{\textbf{x}}        & 0.14          $\pm$ 0      & 71.91          $\pm$ 0.58   & 0.01           $\pm$ 0  & 98.23 $\pm$ 0              & 0.22 $\pm$ 0    & 55.38 $\pm$ 0      & \textbf{0}  $\pm$ 0  & \textbf{100}   $\pm$ 0   \\
 & 80ml, 80cl          & \textcolor[rgb]{0.88,0.13,0.21}{\textbf{x}}               & \textcolor[rgb]{0.88,0.13,0.21}{\textbf{x}}        & 0.14          $\pm$ 0      & 71.36          $\pm$ 0      & 0.01           $\pm$ 0  & 98.23 $\pm$ 0              & 0.22 $\pm$ 0    & 55.69 $\pm$ 0      & \textbf{0}  $\pm$ 0  & \textbf{100}   $\pm$ 0  \\
 & 100ml, 100cl        & \textcolor[rgb]{0.88,0.13,0.21}{\textbf{x}}               & \textcolor[rgb]{0.88,0.13,0.21}{\textbf{x}}        & 0.14          $\pm$ 0      & 71.36          $\pm$ 0      & 0.01           $\pm$ 0  & 98.23 $\pm$ 0              & 0.23 $\pm$ 0    & 53.18 $\pm$ 0      & \textbf{0}  $\pm$ 0  & \textbf{100}   $\pm$ 0  \\
\cmidrule[1.1pt]{1-12}
\end{tabular}}
\end{table*}

\begin{figure*}[!htb]
\begin{center}
\includegraphics[width=0.98\textwidth]{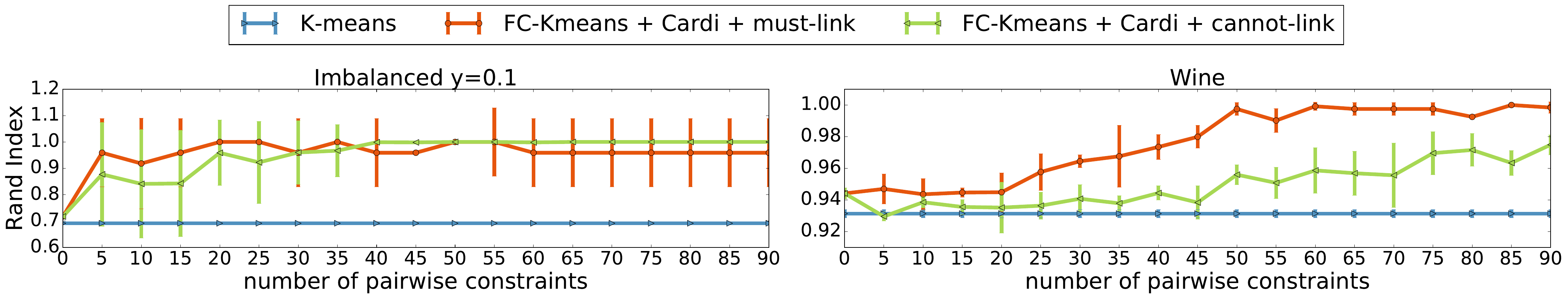}
\end{center}
\caption{Effect of increasing must-link and cannot-link constraints separately, as compared to unconstrained K-means.}
\label{fig:inc_const}
\end{figure*}

\begin{table*}[!htb]
\scriptsize
\centering
\caption{Comparison of several pairwise constrained clustering methods against $\ell_p$Km-Pair using \texttt{\emph{MI}} and \texttt{\emph{HI(\%)}}. The cells indicated with x imply that the underlying method does not attain a feasible solution after 1000 runs.}
\label{table:mix}
\scalebox{0.8}{
\begin{tabular}{|c|cccc|cccc|cccc|}
\cmidrule[1.1pt]{1-13}
 {\multirow{2}{*}{Datasets}} & \multicolumn{4}{c}{$\ell_p$Km-Car} &  \multicolumn{4}{|c}{$\ell_p$Km-Pair}     & \multicolumn{4}{|c|}{$\ell_p$Km-Mix}   \\
 \\[-8pt]
  \cline{2-13}
\\[-7pt]
 &    \multicolumn{1}{c}{ARI($\%$)} & \multicolumn{1}{c}{MI} & \multicolumn{1}{c}{HI($\%$)} & \multicolumn{1}{c|}{Time} & \multicolumn{1}{c}{ARI($\%$)} & \multicolumn{1}{c}{MI} & \multicolumn{1}{c}{HI($\%$)}  &\multicolumn{1}{c|}{Time}  & \multicolumn{1}{c}{ARI($\%$)} & \multicolumn{1}{c}{MI($\%$)}& \multicolumn{1}{c}{HI}  & \multicolumn{1}{c|}{Time}              \\
\\[-8pt]
\hline
  ionosphere            & 8.49  $\pm$ 0     & 0.46           $\pm$ 0             & 9.03            $\pm$ 0        & \textbf{1.50 sec}   & 35.65 $\pm$ 0.67           & 0.321 $\pm$ 0.003    & 35.68 $\pm$ 0.57     & 5.00 sec           & \textbf{80.42}  $\pm$ 4.35    & \textbf{0.097}  $\pm$ 0.002     & \textbf{80.54}  $\pm$ 4.32    & 6.57 sec      \\
  Hepatitis             & 17.12 $\pm$ 1.5   & 0.36           $\pm$ 0.01          & 29.07           $\pm$ 1.28     & \textbf{0.55 sec}   & 38.17 $\pm$ 4.01           & 0.303 $\pm$ 0.021    & 39.49 $\pm$ 4.33     & 0.99 sec           & \textbf{46.25}  $\pm$ 15.74   & \textbf{0.230}  $\pm$ 0.067     & \textbf{54}     $\pm$ 13.47   & 1.09 sec      \\
  Hepatitis1            & 40.98 $\pm$ 7.48  & 0.24           $\pm$ 0.03          & 52.86           $\pm$ 5.29     & 0.91 sec            & 26.97 $\pm$ 12.26          & 0.358 $\pm$ 0.065    & 28.50 $\pm$ 12.96    & \textbf{0.70 sec}  & \textbf{77.27}  $\pm$ 10.01   & \textbf{0.091}  $\pm$ 0.040     & \textbf{81.85}  $\pm$ 7.99    & 1.26 sec   \\
  Breast Cancer Wis-D   & 74.93 $\pm$ 0     & \textbf{0.125} $\pm$ 0             & \textbf{75.03}  $\pm$ 0        & 3.36 sec            & 74.93 $\pm$ 0              & 0.128 $\pm$ 0        & 74.42 $\pm$ 0        & \textbf{3.27 sec}  & \textbf{74.93}  $\pm$ 0       & \textbf{0.125} $\pm$ 0          & \textbf{75.03}  $\pm$ 0       & 9.27 sec    \\
\cmidrule[1.1pt]{1-13}
\end{tabular}}
\end{table*}

\noindent\textbf{(i) Traditional K-means versus $\ell_p$Km.} First, we start by comparing our vanilla constrained free version clustering method $\ell_p$Km against K-means. We show that $\ell_p$Km method can in fact attain very similar,
if not better, performance than traditional K-means (builtin MATLAB function). This is clearly because both methods use the same objective value and that $\ell_p$Km does converge to good solutions. Experiments are conducted on some of the UCI
datasets \cite{bradley2000constrained} (wine, iris and glass). Table \ref{kmeans_oursuncon} reports the K-means objective value, \texttt{\emph{ARI}}(\%), \texttt{\emph{MI}} and \texttt{\emph{HI}}(\%) metrics for both methods.

\noindent\textbf{(ii) Traditional K-means versus $\ell_p$Km-Car.} Here, we demonstrate that our framework coupled with only cardinality constraints outperforms traditional
unconstrained K-means on a variety of synthetic data. 
This highlights the importance of having this prior information available and harnessing it in the clustering process. In these experiments, the cardinality constraints are generated from ground truth labels. To show that cardinality does in fact
help clustering performance, we apply $\ell_p$Km-Car on the two synthetic datasets (\texttt{\emph{Balanced}} and \texttt{\emph{ImBalanced}}) and report their \texttt{\emph{ARI}}, \texttt{\emph{MI}} and \texttt{\emph{HI}} results in Table
\ref{kmeans_card}.

For the \texttt{\emph{Balanced}} dataset, the separation between the four groups of points increases. In fact, K-means tends to cluster points together such that each cluster has a similar variance as other clusters. Consequently, K-means clusters
the high-density points of the \texttt{\emph{Balanced}} dataset together and groups the remaining less dense points into another cluster. In comparison, our framework exploits the cardinality constraints to achieve perfect clustering performance.
Similarly, the \texttt{\emph{ImBalanced}} dataset contains two imbalanced clusters with very different densities, where the separation between them is increased. In this case, K-means often mixes data points between clusters, since the cardinality
constraints are not used. On the other hand, $\ell_p$Km-Car can almost perfectly predict the ground truth clustering labels. Interestingly, the variance of our results is much lower than that of K-means even though they both use the same clustering
initialization. This indicates that the cardinality constraints afford our method robustness to the initialization.

To the best of our knowledge, all previous work that handles generic cardinality constraints do not have readily available code for comparison. Therefore, we only compare our method with traditional unconstrained K-means, so as to demonstrate the effectiveness of adding  cardinality constraints to an unconstrained clustering method.


\noindent\textbf{(iii) Pairwise Constrained Clustering Methods versus $\ell_p$Km-Pair.} Here, we compare our $\ell_p$Km-Pair method against several pairwise constrained methods from the literature, namely Constrained Clustering
\cite{wagstaff2001constrained} (COP-KMEANS), Spectral Clustering \cite{lu2010constrained}, Penalized probabilistic Clustering (PPC) \cite{lu2007penalized} and CCL \cite{dai2003techniques}. All pairwise constraints were randomly generated. Among
these methods, only COP-KMEANS \cite{wagstaff2001constrained} and $\ell_p$Km-Pair exactly enforce the constraints, while the others incorporate them as soft pairwise constraints in their clustering framework. Consequently, Spectral Clustering, PPC
and CCL may
result in clustering violations. However, due to the heuristic nature of COPKMEANS, it may lead to a situation where depending on the initialization no feasible solution is attained. We run all five methods on two  UCI datasets (wine, iris) and
ensure that all methods receive the same randomly generated pairwise constraints. Tables \ref{table:pair_com1} and \ref{table:pair_com2} report the performance of these methods on all discussed metrics. For each experiment, we also report the
number of must-link (ml) and cannot-link (cl) constraints, as well as, the number of must-link violations (MLV) and the cannot-link violations (CLV). It is clear that $\ell_p$Km-Pair outperforms all other methods, while satisfying all the
constraints.


\vspace{3pt}\noindent\textbf{(iv) K-means versus $\ell_p$Km-Mix.} Here, we demonstrate the main motivation behind our flexible framework, namely its ability to incorporate both cardinality and pairwise constraints simultaneously in the clustering
optimization. Firstly, and following previous work \cite{wagstaff2001constrained}, we demonstrate that increasing the number of pairwise constraints (either must-link or cannot-link) with the same cardinality constraints consistently improves
performance. We conduct this experiment on two different datasets: one synthetic ($\texttt{\emph{ImBalanced} y=0.1}$) and one real (wine). Figure \ref{fig:inc_const} compares our method against traditional K-means in such setup. Obviously, K-means
does not benefit from the constraints while ours consistently improves in performance. Secondly, we compare all three variants of our framework, i.e. cardinality only constraints ($\ell_p$Km-Car), pairwise only constraints ($\ell_p$Km-Pair) and
both ($\ell_p$Km-Mix), on several UCI datasets (ionosphere, Hepatitis, Hepatitis1 and Breast Cancer Wis-D). The number of must-link and cannot-link constraints were equal for each dataset and they were set proportional to the dataset size to
(20,25,20,100) respectively. Results in Table \ref{table:mix} show that our method performs increasingly and significantly better, when more constraint categories are used simultaneously. This improvement reaches as high as 40$\%$ in
\texttt{\emph{ARI}} for some datasets. We also report in table \ref{table:mix} the runtime for all 3 different varients on all 4 datasets. The time vary depending on the dataset size and the number of clusters from $0.5-10$ seconds. We did not
compare against other methods here, since there is no existing work that combines both categories of constraints in a unified framework and extending the pairwise constrained methods to cardinality constraints is not trivial.

\section{Conclusion}
We proposed a new flexible framework to handle both pairwise and cardinality constraints for K-means clustering. The resulting integer program is transformed into an equivalent continuous reformulation where pairwise constraints are incorporated as quadratic constraints. The resultant problem is solved using ADMM. Extensive experiments have been conducted on both synthetic and real datasets to demonstrate the competitive performance of our method under different constraint choices and that the proposed method achieves state-of-art performance when both types of constraints are used simultaneously. As a future work, we seek to adapt our work to deep clustering approaches \cite{alqahtani2018deep,alqahtani2019learning, alqahtani2021deep}, which jointly learn feature representations and cluster assignments. That is to say, the ADMM updates will involove another minimization step that trains the encoder for the feature learning procedure while the other steps involve on sloving both the assignment and imposing the constraints. This shall have applications in biology, medicine, finance, and animation, as in many such applications, where pseudo labales in terms of assignments are provided as constraints, e.g., two patients are labeled to have a similar syndrome, feature learning will allow for further improvements in clustering performance.

\section*{Acknowledgment}
The work was done when Adel Bibi was at KAUST. This work was supported by the King Abdullah University of Science and Technology (KAUST) Office of Sponsored Research and the  Deanship  of  Scientific  Research,  King  Khalid  University  of  Kingdom  of  Saudi  Arabia  under research grant  number  (RGP1/357/43).

\begin{IEEEbiography}[{\includegraphics[width=1in,height=1.25in,clip,keepaspectratio]{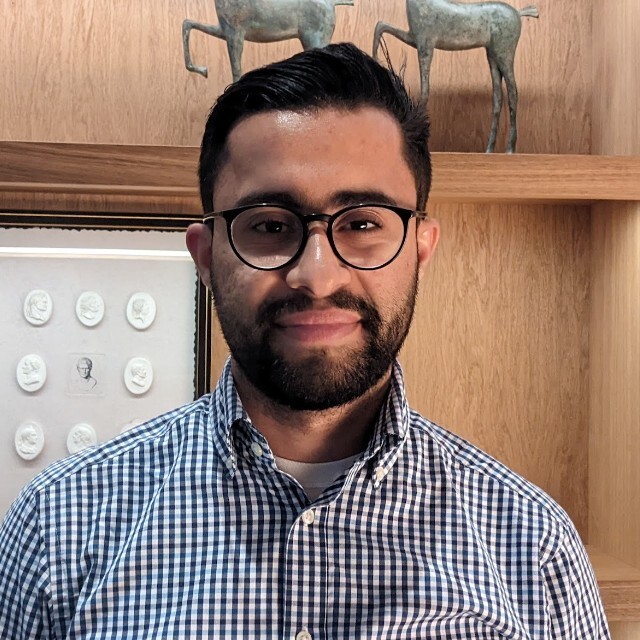}}]{Adel Bibi} is a senior research fellow in machine learning and computer vision at the Department of Engineering Science of the University of Oxford. He is also a Junior Research Fellow (JRF) of Kellogg College. Prior to that, he was a postdoctoral researcher at the same group for a year since October 2020. Adel received his MSc and PhD degrees from King Abdullah University of Science and Technology (KAUST) in 2016 and 2020, respectively. He received his BSc degree in electrical engineering with class honors from Kuwait university in 2014. Adel has been recognized as an outstanding reviewer for CVPR18, CVPR19, ICCV19, and ICLR22, and won the best paper award at the optimization and big data conference in KAUST. He has published more than 20 papers in CVPRs, ECCVs, ICCVs and ICLRs some which were selected as orals and spotlights and is going to serve as an area chair for AAAI23.
\end{IEEEbiography}

\begin{IEEEbiography}[{\includegraphics[width=1in,height=1.25in,keepaspectratio]{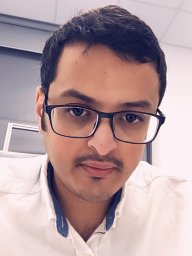}}]{Ali AlQahtani} received his Ph.D. in computer science from Swansea University, Swansea, U.K., in 2021. He is currently an Assistant Professor with the Department of Computer Science, King Khalid University, Abha, KSA. He has published several refereed conference and journal publications. His research interests include various aspects of pattern recognition, deep learning, and machine intelligence and their applications to real-world problems.
\end{IEEEbiography}

\begin{IEEEbiography}[{\includegraphics[width=1in,height=1.25in,clip,keepaspectratio]{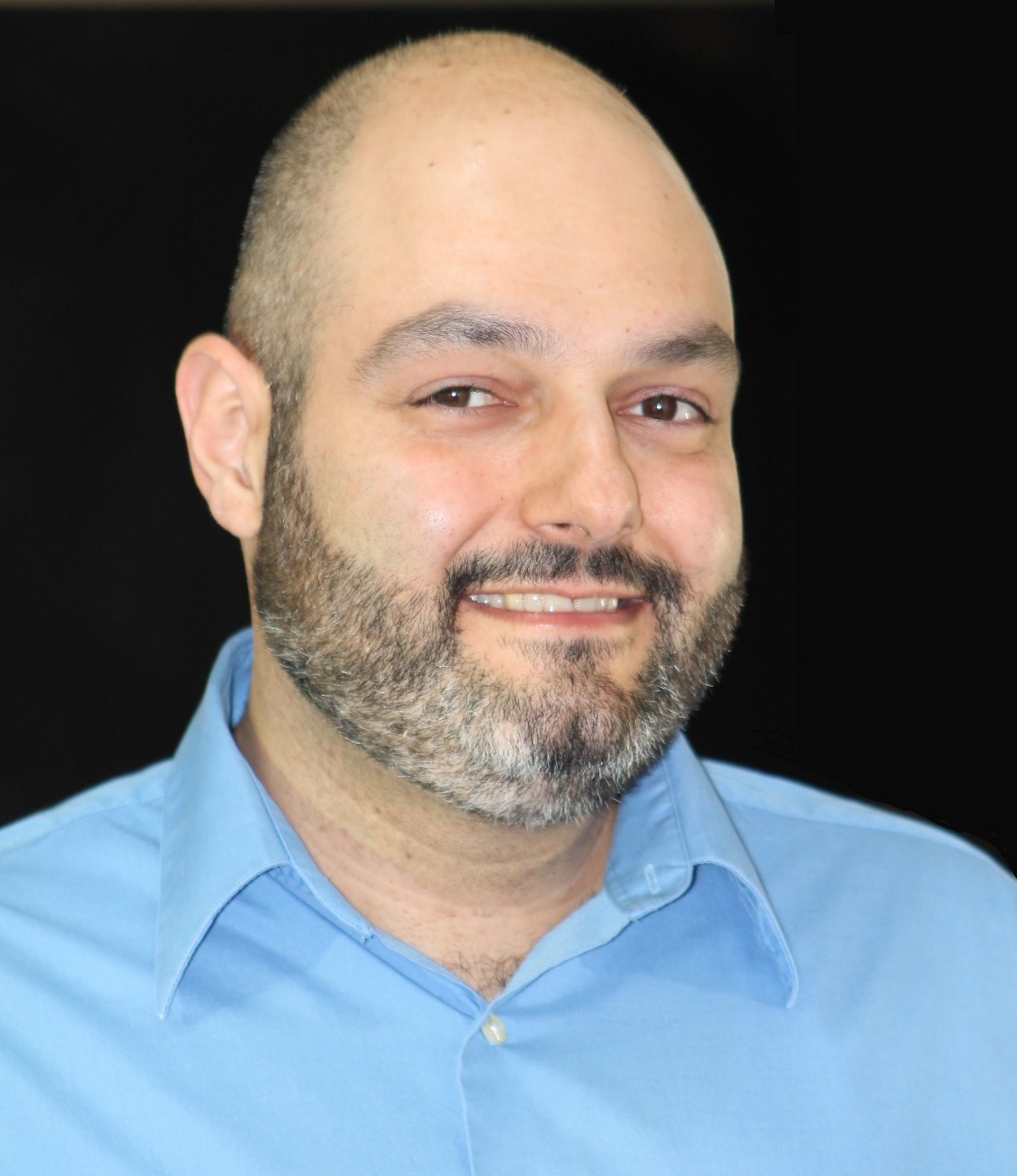}}]{Bernard Ghanem} is currently a Professor in the Electrical and Computer Engineering program at King Abdullah University of Science and Technology (KAUST). His research interests lie in computer vision and machine learning with emphasis on topics in video understanding, 3D recognition, and theoretical foundations of deep learning. He received his Bachelor's degree from the American University of Beirut (AUB) in 2005 and his MS/PhD from the University of Illinois at Urbana-Champaign (UIUC) in 2010. His work has received several awards and honors, including several Best Workshop Paper Awards in CVPR and ECCV, a Google Faculty Research Award in 2015 (1st in MENA for Machine Perception), and a Abdul Hameed Shoman Arab Researchers Award for Big Data and Machine Learning in 2020. He has co-authored more than 100 peer reviewed conference and journal papers in his field as well as three issued patents. He serves as Associate Editor for IEEE Transactions on Pattern Analysis and Machine Intelligence (TPAMI) and has served as Area Chair (AC) for the leading computer vision and machine learning conferences.
\end{IEEEbiography}

\section{Appendix}
\section*{Reformulation of Constrained K-means.} 

We start with our formulation of constrained K-means in Equation (1) below (or Equation (4) in the manuscript).
\begin{align*}
    &\min_{\mathbf{x},\mathbf{w},\mathbf{z}_1,\mathbf{z}_2,\mathbf{z}_3, \mathbf{z}_4} \sum_{j=1}^k \sum_{i=1}^n x_{ij} \left\| \mathbf{s}_i - \mathbf{S} \Lambda_j\mathbf{w}\right\|_2^2 \\
    & \text{s.t.} \Psi^{\top} \mathbf{x} = \mathbf{1}_n, \mathbf{z}_1 \in S_b, \mathbf{z}_2 \in S_2, \mathbf{Q}\mathbf{P}^\top \mathbf{x} = \mathbf{u}, \\
    & \mathbf{x} = \mathbf{P}\mathbf{w} \odot \mathbf{C}\mathbf{x}, \left(\mathbf{E}_1\mathbf{z}_3\right)^\top \mathbf{E}_2 \mathbf{x} = v, \left(\mathbf{E}_3\mathbf{z}_4\right)^\top \mathbf{E}_4 \mathbf{x} = 0, \\
    & \mathbf{x} = \mathbf{z}_1, \mathbf{x} = \mathbf{z}_2, \mathbf{x} = \mathbf{z}_3, \mathbf{x} = \mathbf{z}_4.
\end{align*}

We define $\mathbf{X} \in \mathbb{R}^{k\times n}$ and therefore $\mathbf{x} = \text{vect}(\mathbf{X})$. The
vect operator simply vectorizes the matrix one column at a time (i.e. one data point at a time). The order of concatenation is reverse for the label vector: $\mathbf{w} = \text{vect}(\mathbf{W})$, where
$\mathbf{W} \in \mathbb{R}^{n \times k}$ (i.e. the vectorization is done one cluster at a
time). Therefore, it is important to point out that the order of the binary labels $\mathbf{x}$ and that of $\mathbf{w}$ is swapped. Reordering these vectors based on data points or clusters is controlled by the permutation matrix $\mathbf{P} \in \mathbb{R}^{nk \times nk}$. Therefore, $\mathbf{P}\mathbf{x} = \mathbf{P}^\top \mathbf{x} = \text{vect}(\mathbf{X}^\top)$ and $\mathbf{P}\mathbf{w} = \mathbf{P^\top}\mathbf{w}= \text{vect}(\mathbf{W}^\top)$. Throughout the main manuscript and the appendix, we use $\mathbf{P}$ and $\mathbf{P}^\top$ to change the order of w to the same
order as $\mathbf{x}$ and vice versa. Matrix $\mathbf{S} \in \mathbb{R}^{d \times n}$ in Equation (1)
simply concatenates all the data points in its columns, where $\Lambda_j \in \mathbb{R}^{n \times nk}$ is zero everywhere except for the $\text{j}^{\text{th}}$ block that is identity, i.e. $\Lambda_j = \begin{bmatrix} = \mathbf{0}, \dots, \mathbf{I}_k^j, \dots  \end{bmatrix}$.

As for $\Psi^{\top} \in \mathbb{R}^{n \times n k}$, it is a binary matrix that has in each row a vector $\mathbf{1}_{k}^{\top}$ that sums all the binary labels for a given data point while the rest are zeros. We write this matrix in blockwise form as follows:

\begin{align*}
\Psi^{\top}=
\begin{bmatrix}
\mathbf{1}_{k}^{\top} & \mathbf{0}_{k} & \cdots \\
\mathbf{0}_{k}^{\top} & \mathbf{1}_{k}^{\top} & \mathbf{0}_{k} \\
\vdots & \vdots & \vdots \\
\mathbf{0}_{k} & \cdots & \mathbf{1}_{k}
\end{bmatrix}
\end{align*}

As for $\mathbf{Q} \in \mathbb{R}^{k \times n k}$, it sums all the binary labels of each cluster at a time for all the data points and its blockwise matrix form is given as follows:

\begin{align*}
\mathbf{Q}=
\begin{bmatrix}
\mathbf{1}_{n}^{T} & \mathbf{0}_{n k-n}^{T} & \ldots & \\
\mathbf{0}_{n}^{T} & \mathbf{1}_{n}^{T} & \mathbf{0}_{n k-n}^{T} & \cdots \\
& \cdots & \cdots & &
\end{bmatrix}
\end{align*}

As for $\mathbf{C} \in \mathbb{R}^{n k \times n k}$, it sums the binary labels for each cluster at a time and its blockwise matrix form is as follows:

\begin{align*}
\mathbf{C}=
\begin{bmatrix}
\mathbf{I}_{k} & \cdots & \mathbf{I}_{k} \\
\vdots & \cdots & \vdots \\
\mathbf{I}_{k} & \cdots & \mathbf{I}_{k}
\end{bmatrix}
\end{align*}

As for the box and $\ell_{2}$-sphere constraints (which intersects with the binary vector space), we define two sets: $S_{b}:=$ $\{\mathbf{a}: \mathbf{0} \leq \mathbf{a} \leq \mathbf{1}\}$ and $S_{2}:=\left\{\mathbf{a} \in \mathbb{R}^{n}:\left\|\mathbf{a}-\frac{1}{2} \mathbf{1}\right\|_{2}^{2}=\frac{n}{2}\right\}$, respectively. It is shown in \cite{Wu2016ell_pBoxAA} that: $\{0,1\}^{n}=S_{b} \cap S_{2}$.

Lastly, $\mathbf{E}_{1}, \mathbf{E}_{2} \in \mathbb{R}^{k v \times n k}$ and $\mathbf{E}_{3}, \mathbf{E}_{4} \in \mathbb{R}^{k e \times n k}$ are selection matrices for the must-link and cannot-link constraints respectively. They select the data points that are involved in both types of constraints.

\textbf{Applying ADMM.} Following the conventional treatment of an optimization problem using ADMM, we first formulate the augmented Lagrangian function is given as follows:

\begin{align*}
&\mathcal{L}\left(\mathbf{x}, \mathbf{w}, \mathbf{z}_{1}, \mathbf{z}_{2}, \mathbf{z}_{3}, \mathbf{z}_{4}, \mathbf{y}_{1}, \mathbf{y}_{2}, \mathbf{y}_{3}, \mathbf{y}_{4}, \mathbf{y}_{5}, y_{6}, \mathbf{y}_{7}, y_{8}, \mathbf{y}_{9}\right):= \\
&\sum_{j=1}^{k} \sum_{i=1}^{n} x_{i j}\left\|\mathbf{s}_{i}-\mathbf{S} \Lambda_{j} \mathbf{w}\right\|_{2}^{2}+\mathbf{y}_{1}^{\top}\left(\Psi^{\top} \mathbf{x}-\mathbf{1}_{n}\right)+\frac{\rho_{1}}{2}\left\|\Psi^{\top} \mathbf{x}-\mathbf{1}_{n}\right\|_{2}^{2} + \\
&\mathbb{I}_{\left\{\mathbf{z}_{1} \in S_{b}\right\}}+\mathbf{y}_{2}^{\top}\left(\mathbf{x}-\mathbf{z}_{1}\right)+\frac{\rho_{2}}{2}\left\|\mathbf{x}-\mathbf{z}_{1}\right\|_{2}^{2}+\mathbb{I}_{\left\{\mathbf{z}_{2} \in S_{2}\right\}}+\mathbf{y}_{3}^{\top}\left(\mathbf{x}-\mathbf{z}_{2}\right)+ \\
& \frac{\rho_{3}}{2}\left\|\mathbf{x}-\mathbf{z}_{2}\right\|_{2}^{2}+\mathbf{y}_{4}^{\top}\left(\mathbf{Q} \mathbf{P}^{\top} \mathbf{x}-\mathbf{u}\right)+\frac{\rho_{4}}{2}\left\|\mathbf{Q P}^{\top} \mathbf{x}-\mathbf{u}\right\|_{2}^{2}+ \\
& \mathbf{y}_{5}^{\top}\left(\mathbf{I}-\operatorname{diag}(\mathbf{P w}) \mathbf{C}\right) \mathbf{x}+\frac{\rho_{5}}{2}\|\left(\mathbf{I}-\operatorname{diag}(\mathbf{P w}) \mathbf{C}\right) \mathbf{x}\|_{2}^{2}+ \\
&y_{6}\left(\mathbf{z}_{3}^{\top} \mathbf{E}_{1}^{\top} \mathbf{E}_{2} \mathbf{x}-v\right)+\frac{\rho_{6}}{2}\left\|\mathbf{z}_{3}^{\top} \mathbf{E}_{1}^{\top} \mathbf{E}_{2} \mathbf{x}-v\right\|_{2}^{2}+\mathbf{y}_{7}^{\top}\left(\mathbf{x}-\mathbf{z}_{3}\right)+ \\
& \frac{\rho_{7}}{2}\left\|\mathbf{x}-\mathbf{z}_{3}\right\|_{2}^{2}+y_{8}\left(\mathbf{z}_{4}^{\top} \mathbf{E}_{3}^{\top} \mathbf{E}_{4} \mathbf{x}\right)+\frac{\rho_{8}}{2}\left\|\mathbf{z}_{4}^{\top} \mathbf{E}_{3}^{\top} \mathbf{E}_{4} \mathbf{x}\right\|_{2}^{2} + \\
&\mathbf{y}_{9}^{\top}\left(\mathbf{x}-\mathbf{z}_{4}\right)+\frac{\rho_{9}}{2}\left\|\mathbf{x}-\mathbf{z}_{4}\right\|_{2}^{2}.
\end{align*}

ADMM updates steps tend to update each primal variable ( $\mathbf{x}, \mathbf{w}$, and $\mathbf{z}_{1-4}$ sequentially, while keeping the rest of these variables and the dual variables $\mathbf{y}_{1-5}, \mathbf{y}_{6,8}$, and $\mathbf{y}_{6,9})$ set to their most recent values. After the primal variables are updated, the dual variables are updated via a single gradient ascent step. Next, we detail each update step and the underlying optimization sub-problem that needs to be solved.

\textbf{Update $\mathbf{x}$:}  

\begin{align*}
&\mathbf{x} \leftarrow \underset{\mathbf{x}}{\arg \min } \sum_{j=1}^{k} \sum_{i=1}^{n} x_{i j}\left\|\mathbf{s}_{\mathbf{i}}-\mathbf{S} \Lambda_{j} \mathbf{w}\right\|_{2}^{2}+\mathbf{y}_{1}^{\top}\left(\Psi^{\top} \mathbf{x}-\mathbf{1}_{n}\right)+ \\
&\frac{\rho_{1}}{2}\left\|\Psi^{\top} \mathbf{x}-\mathbf{1}_{n}\right\|_{2}^{2}+\mathbf{y}_{2}^{\top}\left(\mathbf{x}-\mathbf{z}_{1}\right)+\frac{\rho_{2}}{2}\left\|\mathbf{x}-\mathbf{z}_{1}\right\|_{2}^{2}+\mathbf{y}_{3}^{\top}\left(\mathbf{x}-\mathbf{z}_{2}\right)+ \\
& \frac{\rho_{3}}{2}\left\|\mathbf{x}-\mathbf{z}_{2}\right\|_{2}^{2}+\mathbf{y}_{4}^{\top}\left(\mathbf{Q} \mathbf{P}^{\top} \mathbf{x}-\mathbf{u}\right)+\frac{\rho_{4}}{2}\left\|\mathbf{Q} \mathbf{P}^{\top} \mathbf{x}-\mathbf{u}\right\|_{2}^{2}+ \\
& \mathbf{y}_{5}^{\top}(\mathbf{I}-\operatorname{diag}(\mathbf{P} \mathbf{w}) \mathbf{C}) \mathbf{x}+\frac{\rho_{5}}{2}\|(\mathbf{I}-\operatorname{diag}(\mathbf{P} \mathbf{w}) \mathbf{C}) \mathbf{x}\|_{2}^{2}+ \\
& y_{6}\left(\mathbf{z}_{4}^{\top} \mathbf{E}_{1}^{\top} \mathbf{E}_{2} \mathbf{x}-v\right)+\frac{\rho_{6}}{2}\left\|\mathbf{z}_{4}^{\top} \mathbf{E}_{1}^{\top} \mathbf{E}_{2} \mathbf{x}-v\right\|_{2}^{2}+\mathbf{y}_{7}^{\top}\left(\mathbf{x}-\mathbf{z}_{4}\right)+ \\
& \frac{\rho_{7}}{2}\left\|\mathbf{x}-\mathbf{z}_{4}\right\|_{2}^{2}+y_{8}\left(\mathbf{z}_{5}^{\top} \mathbf{E}_{3}^{\top} \mathbf{E}_{4} \mathbf{x}\right)+\frac{\rho_{8}}{2}\left\|\mathbf{z}_{5}^{\top} \mathbf{E}_{3}^{\top} \mathbf{E}_{4} \mathbf{x}\right\|_{2}^{2}+\mathbf{y}_{9}^{\top}\left(\mathbf{x}-\mathbf{z}_{5}\right)+ \\
&\frac{\rho_9}{2}\left\|\mathbf{x}-\mathbf{z}_{5}\right\|_{2}^{2}
\end{align*}

The aforemention problem is strongly convex quadratic in $\mathbf{x}$. Therefore, a stationary point is necessary and sufficient for optimality. By equating the gradient to zero, we get:

\begin{align*}
&\Big(\rho_{1} \Psi \Psi^{\top}+\left(\rho_{2}+\rho_{3}+\rho_{7}+\rho_{9}\right) \mathbf{I}_{n k}+\rho_{4} \mathbf{P} \mathbf{Q}^{\top} \mathbf{Q} \mathbf{P}^{\top}+ \\
&\rho_{5}(\mathbf{I}-\operatorname{diag}(\mathbf{P} \mathbf{w}) \mathbf{C})^{\top}(\mathbf{I}-\operatorname{diag}(\mathbf{P} \mathbf{w}) \mathbf{C})+ \\
&\rho_{6} \mathbf{E}_{2}^{\top} \mathbf{E}_{1} \mathbf{z}_{3} \mathbf{z}_{3}^{\top} \mathbf{E}_{1}^{\top} \mathbf{E}_{2}+\rho_{8} \mathbf{E}_{4}^{\top} \mathbf{E}_{4} \mathbf{z}_{4} \mathbf{z}_{5}^{\top} \mathbf{E}_{3}^{\top} \mathbf{E}_{4}\Big) \mathbf{x}= \\
&-\left(\operatorname{vect}(\mathbf{B})+\Psi \mathbf{y}_{1}+\mathbf{y}_{2}+\mathbf{y}_{3}-\rho_{1} \Psi \mathbf{1}_{n}-\rho_{2} \mathbf{z}_{1}-\right. \\
&\rho_{3} \mathbf{z}_{2}-\mathbf{C}^{\top} \operatorname{diag}(\mathbf{P} \mathbf{w}) \mathbf{y}_{5}+y_{6} \mathbf{E}_{2}^{\top} \mathbf{E}_{1} \mathbf{z}_{4}-\rho_{6} v \mathbf{E}_{2}^{\top} \mathbf{E}_{1} \mathbf{z}_{3}+\mathbf{y}_{7}- \\
&\left.\rho_{7} \mathbf{z}_{3}+y_{8} \mathbf{E}_{4}^{\top} \mathbf{E}_{3} \mathbf{z}_{4}+\mathbf{y}_{9}-\rho_{9} \mathbf{z}_{4}+\mathbf{P} \mathbf{Q}^{\top} \mathbf{y}_{4}-\rho_{4} \mathbf{P} \mathbf{Q}^{\top} \mathbf{u}\right),
\end{align*}
where $\mathbf{B}(i, j)=\left\|\mathbf{s}_{\mathbf{i}}-\mathbf{S} \Lambda_{j} \mathbf{w}\right\|_{2}^{2}$.

\textbf{Update $\mathbf{w}$:}
\begin{align*}
&\mathbf{w} \leftarrow \underset{\mathbf{w}}{\arg \min } \sum_{j=1}^{k} \sum_{i=1}^{n} x_{i j}\left\|\mathbf{s}_{\mathbf{i}}-\mathbf{S} \Lambda_{j} \mathbf{w}\right\|_{2}^{2}+ \\
&\mathbf{y}_{5}^{T}(\mathbf{I}-\operatorname{diag}(\mathbf{P} \mathbf{w}) \mathbf{C}) \mathbf{x}+\frac{\rho_{5}}{2}\left\|(\mathbf{I}-\operatorname{diag}(\mathbf{P} \mathbf{w}) \mathbf{C}) \mathbf{x}\right\|_{2}^{2}.
\end{align*}
\noindent Similarly to the $\mathrm{x}$-update, the problem is strongly convex quadratic and finding a stationary point is necessary and sufficient for a global solution. Thus, the gradient is given by:

\begin{align*}
&-\sum_{j=1}^{k} \sum_{i=1}^{n} 2 x_{i j}\left(\mathbf{S} \Lambda_{j}\right)^{T}\left(\mathbf{s}_{i}-\mathbf{S} \Lambda_{j} \mathbf{w}\right)-\mathbf{P}^{T} \mathbf{C} \mathbf{x} \odot \mathbf{y}_{5} - \\
&\rho_{5}\left(\mathbf{P}^{T} \operatorname{diag}(\mathbf{C} \mathbf{x})\right)(\mathbf{I}-\operatorname{diag}(\mathbf{P} \mathbf{w}) \mathbf{C}) \mathbf{x}=0.
\end{align*}

Then, we have
\begin{align*}
&-\sum_{j=1}^{k} \sum_{i=1}^{n} 2 x_{i j} \Lambda_{j}^{T} \mathbf{S}^{T} \mathbf{s}_{i}+\sum_{j=1}^{k} \sum_{i=1}^{n} 2 x_{i j} \Lambda_{j}^{T} \mathbf{S}^{T} \mathbf{S} \Lambda_{j} \mathbf{w}-\mathbf{P}^{T} \mathbf{C x} \odot \mathbf{y}_{5} \\
&-\rho_{5} \mathbf{P}^{T} \mathbf{C} \mathbf{x} \odot \mathbf{x}+\rho_{5} \mathbf{P}^{T} \operatorname{diag}(\mathbf{C x}) \operatorname{diag}(\mathbf{P} \mathbf{w}) \mathbf{C} \mathbf{x}=0.
\end{align*}

Therefore,
\begin{align*}
&\left[\sum_{j=1}^{k} \sum_{i=1}^{n} 2 x_{i j} \Lambda_{j}^{T} \mathbf{S}^{T} \mathbf{S} \Lambda_{j} \mathbf{w}+\rho_{5} \mathbf{P}^{T} \operatorname{diag}(\mathbf{C} \mathbf{x}) \operatorname{diag}(\mathbf{P} \mathbf{w})\mathbf{C}\mathbf{x} \right] \\
&=\sum_{j}^{k} \sum_{i}^{n} 2 x_{i j} \Lambda_{j}^{T} \mathbf{S}^{T} \mathbf{s}_{i}+\mathbf{P}^{T} \mathbf{C} \mathbf{x} \odot \mathbf{y}_{5}+\rho_{5} \mathbf{P}^{T} \mathbf{C x} \odot \mathbf{x}
\end{align*}

Finally, we have,
\begin{align*}
&{\left[\sum_{j=1}^{k} \sum_{i=1}^{n} 2 x_{i j} \Lambda_{j}^{T} \mathbf{S}^{T} \mathbf{S} \Lambda_{j}+\rho_{5} \mathbf{P}^{T} \operatorname{diag}(\mathbf{C} \mathbf{x} \odot \mathbf{C} \mathbf{x}) \mathbf{P}\right] \mathbf{w}} \\
&=\sum_{j=1}^{k} \sum_{i=1}^{n} 2 x_{i j} \Lambda_{j}^{T} \mathbf{S}^{T} \mathbf{s}_{i}+\mathbf{P}^{T} \mathbf{C} \mathbf{x} \odot \mathbf{y}_{5}+\rho_{5} \mathbf{P}^{T} \mathbf{C} \mathbf{x} \odot \mathbf{x}
\end{align*}

In this derivation, we use the fact that $\nabla_{\mathbf{w}}\left(\mathbf{y}_{5}^{T} \mathbf{P} \mathbf{w} \odot\right.$ $\mathbf{C x})=\mathbf{P}^{T} \mathbf{C} \mathbf{x} \odot \mathbf{y}_{5}$. Tp see why this is the case, note the following identities: $\mathbf{a} \odot \mathbf{b}=\mathbf{b} \odot \mathbf{a}=\operatorname{diag}(\mathbf{a}) \mathbf{b}=$ $\operatorname{diag}(\mathbf{b}) \mathbf{a}$. 

Therefore,
\begin{align*}
\nabla_{\mathbf{w}}\left(\mathbf{y}_{5}^{T} \mathbf{P} \mathbf{w} \odot \mathbf{C} \mathbf{x}\right) &=\nabla_{\mathbf{w}}\left(\mathbf{y}_{5}^{T} \operatorname{diag}(\mathbf{C} \mathbf{x}) \mathbf{P} \mathbf{w}\right) \\
&=\mathbf{P}^{T} \operatorname{diag}(\mathbf{C} \mathbf{x}) \mathbf{y}_{5} \\
&=\mathbf{P}^{T} \mathbf{C} \mathbf{x} \odot \mathbf{y}_{5}
\end{align*}

\textbf{Update $\mathbf{z}_{1}$:}
\begin{align*}
&\mathbf{z}_{1} \leftarrow \underset{\mathbf{z}_{1} \in S_{b}}{\arg \min } \mathbf{y}_{2}^{\top}\left(\mathbf{x}-\mathbf{z}_{1}\right)+\frac{\rho_{2}}{2}\left\|\mathbf{x}-\mathbf{z}_{1}\right\|_{2}^{2} \\
&\mathbf{z}_{1} \leftarrow \underset{\mathbf{z}_{1} \in S_{b}}{\arg \min }\left\|\mathbf{z}_{1}-\left(\mathbf{x}+\frac{\mathbf{y}_{2}}{\rho_{2}}\right)\right\|_{2}^{2} \\
&\mathbf{z}_{1}=\mathbf{P}_{S_{b}}\left(\mathbf{x}+\frac{\mathbf{y}_{2}}{\rho_{2}}\right)
\end{align*}

Here, we need to perform a simple projection onto the box set $S_{b}$. The projection $\mathbf{P}_{S_{b}}(.)$ is an elementwise clamping between 0 and $+1$ that is, $\mathbf{P}_{S_{b}}(a)=\min (\max (a, 0), 1)$ for a scalar value $a$.

\textbf{Update $\mathbf{z}_{2}$:}

\begin{align*}
&\mathbf{z}_{2} \leftarrow \underset{\mathbf{z}_{2} \in S_{2}}{\arg \min } \mathbf{y}_{3}^{\top}\left(\mathbf{x}-\mathbf{z}_{2}\right)+\frac{\rho_{3}}{2}\left\|\mathbf{x}-\mathbf{z}_{2}\right\|_{2}^{2} \\
&\mathbf{z}_{2} \leftarrow \underset{\mathbf{z}_{2} \in S_{2}}{\arg \min }\left\|\mathbf{z}_{2}-\left(\mathbf{x}+\frac{\mathbf{y}_{3}}{\rho_{3}}\right)\right\|_{2}^{2} \\
&\mathbf{z}_{2}=\mathbf{P}_{S_{2}}\left(\mathbf{x}+\frac{\mathbf{y}_{3}}{\rho_{3}}\right)
\end{align*}

\noindent We need to perform a simple projection onto the $\ell_{2^{-}}$ sphere: $S_{2}=\left\{\mathbf{a} \in \mathbb{R}^{n}:\left\|\mathbf{a}-\frac{1}{2} \mathbf{1}\right\|_{2}^{2}=\frac{n}{4}\right\}$. The projection $\mathbf{P}_{\mathcal{S}_{2}}(.)$ (.) involves an elementwise shift and $\ell_{2}$ vector normalization and thus is given as $\mathbf{P}_{\mathcal{S}_{2}}(\mathbf{a})=\frac{\sqrt{n}}{2}\left(\frac{\mathbf{a}-\frac{1}{2} \mathbf{1}}{\left\|\mathbf{a}-\frac{1}{2} \mathbf{1}\right\|_{2}}\right)+\frac{1}{2} \mathbf{1}$, for any vector $\mathbf{a} \in \mathbb{R}^{n}$.

\textbf{Update $\mathbf{z}_{3}$:}
\begin{align*}
&\mathbf{z}_{3} \leftarrow \underset{\mathbf{z}_{3}}{\text{argmin}} ~~ y_{6} (\mathbf{z}_{3}^{\top} \mathbf{E}_{1}^{\top} \mathbf{E}_{2} \mathbf{x}) + \frac{\rho_{6}}{2}\left\|\mathbf{z}_{3}^{\top} \mathbf{E}_{1}^{\top} \mathbf{E}_{2} \mathbf{x}-v\right\|_{2}^{2}-\mathbf{y}_{7}^{\top} \mathbf{z}_{3}+ \\
& \frac{\rho_{7}}{2}\left\|\mathbf{x}-\mathbf{z}_{3}\right\|_{2}^{2}
\end{align*}

\noindent The problem is strongly convex quadratic in $\mathbf{z}_{4}$, so we obtain the unique global minimizer by equating the gradient to zero as follows

\begin{align*}
&{\left[\rho_{6} \mathbf{E}_{1}^{\top} \mathbf{E}_{2} \mathbf{x} \mathbf{x}^{\top} \mathbf{E}_{2}^{\top} \mathbf{E}_{1}+\rho_{7} \mathbf{I}_{n k}\right] \mathbf{z}_{4}=\mathbf{y}_{7}+\rho_{7} \mathbf{x}-y_{6} \mathbf{E}_{1}^{\top} \mathbf{E}_{2} \mathbf{x +}} \\
&\rho_{6} v \mathbf{E}_{1}^{\top} \mathbf{E}_{2} \mathbf{x}
\end{align*}

\textbf{Update $\mathbf{z}_{4}$:}

\begin{align*}
&\mathbf{z}_{4} \leftarrow \underset{\mathbf{z}_{4}}{\text{argmin}} ~~ y_{8} (\mathbf{z}_{4}^{\top} \mathbf{E}_{3}^{\top} \mathbf{E}_{4} \mathbf{x})+\frac{\rho_{8}}{2}\left\|\mathbf{z}_{4}^\top \mathbf{E}_{3}^{\top} \mathbf{E}_{4} \mathbf{x}\right\|_{2}^{2}-\mathbf{y}_{9}^{\top} \mathbf{z}_{4}+ \\
&\frac{\rho_{9}}{2}\left\|\mathbf{x}-\mathbf{z}_{4}\right\|_{2}^{2}
\end{align*}

\noindent The problem is also strongly convex quadratic in $\mathbf{z}_{5}$, so we obtain the unique global minimizer by equating the gradient to zero as follows

\begin{align*}
\left[\rho_{8} \mathbf{E}_{3}^{\top} \mathbf{E}_{4} \mathbf{x} \mathbf{x}^{\top} \mathbf{E}_{4}^{\top} \mathbf{E}_{3}+\rho_{9} \mathbf{I}_{n k}\right] \mathbf{z}_{5}=\mathbf{y}_{9}+\rho_{9} \mathbf{x}-y_{8} \mathbf{E}_{3}^{\top} \mathbf{E}_{4} \mathbf{x}
\end{align*}

\textbf{Update $\mathbf{y}_{1}, \mathbf{y}_{2}, \mathbf{y}_{3}, \mathbf{y}_{4}, \mathbf{y}_{5}, y_{6}, \mathbf{y}_{7}, y_{8}, \mathbf{y}_{9}$:} Lastly, we need to perform dual gradient ascent to update the dual variables as follows:

\begin{align*}
&\mathbf{y}_{1} \leftarrow \mathbf{y}_{1}+\rho_{1}\left(\Psi^{\top} \mathbf{x}-\mathbf{1}_{n}\right), \quad \mathbf{y}_{2} \leftarrow \mathbf{y}_{2}+\rho_{2}\left(\mathbf{x}-\mathbf{z}_{2}\right) \\
&\mathbf{y}_{3} \leftarrow \mathbf{y}_{3}+\rho_{3}\left(\mathbf{x}-\mathbf{z}_{3}\right), \quad \mathbf{y}_{4} \leftarrow \mathbf{y}_{4}+\rho_{4}\left(\mathbf{Q P}^{\top} \mathbf{x}-\mathbf{u}\right) \\
&\mathbf{y}_{5} \leftarrow \mathbf{y}_{5}+\rho_{5}(\mathbf{x}-\mathbf{P w} \odot \mathbf{C x}), \quad y_{6} \leftarrow y_{6}+\rho_{6}\left(\mathbf{z}_{3}^{\top} \mathbf{E}_{1}^{\top}\right. \\
&\mathbf{y}_{7} \leftarrow \mathbf{y}_{7}+\rho_{7}\left(\mathbf{x}-\mathbf{z}_{3}\right), \quad y_{8} \leftarrow y_{8}+\rho_{8}\left(\mathbf{z}_{4}^{\top} \mathbf{E}_{3}^{\top} \mathbf{E}_{4} \mathbf{x}\right) \\
&\mathbf{y}_{9} \leftarrow \mathbf{y}_{9}+\rho_{9}\left(\mathbf{x}-\mathbf{z}_{4}\right)
\end{align*}

\begin{figure}[!htp]
\begin{center}
\includegraphics[width=0.35\textwidth]{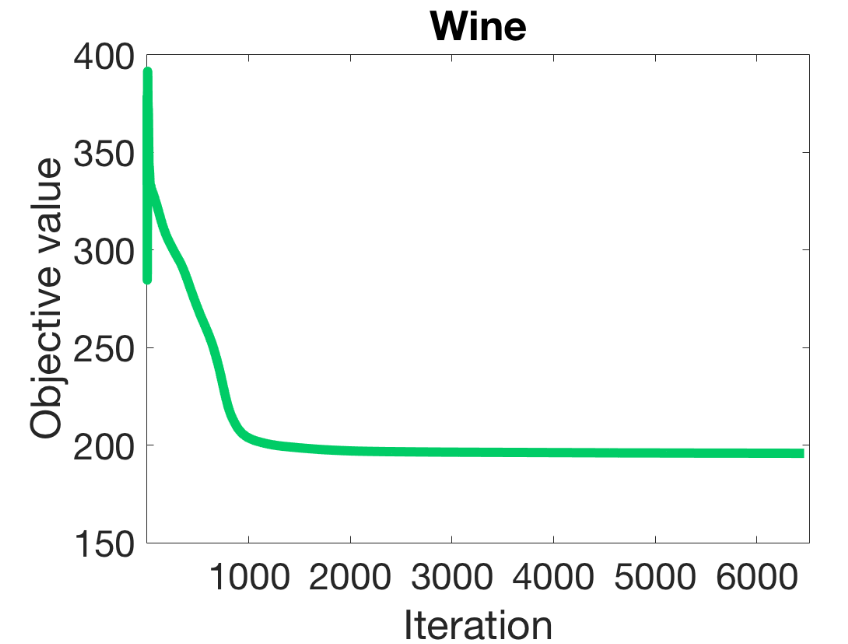}
\caption{Convergence of the K-means objective value us-
ing FCKm with random initialization on the Wine dataset. Note the decreasing nature of the objective and its smooth convergence to the solution}\label{fig:conv_1_appendix}
\end{center}
\end{figure}

\begin{figure}[t]
\begin{center}
\includegraphics[width=0.5\textwidth]{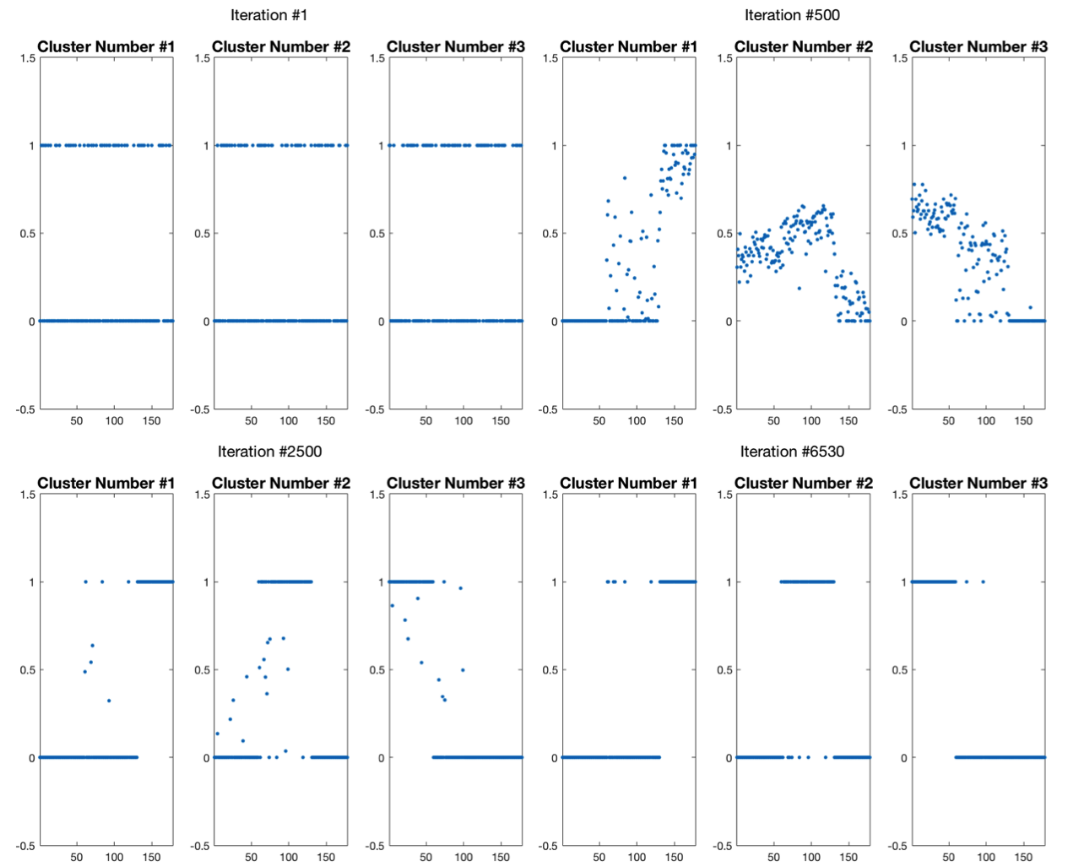}
\caption{Convergence of the solution x using FCKm with random initialization on the Wine dataset.}\label{fig:conv_2_appendix}
\end{center}
\end{figure}

\begin{figure}[t]
\begin{center}
\includegraphics[width=0.35\textwidth]{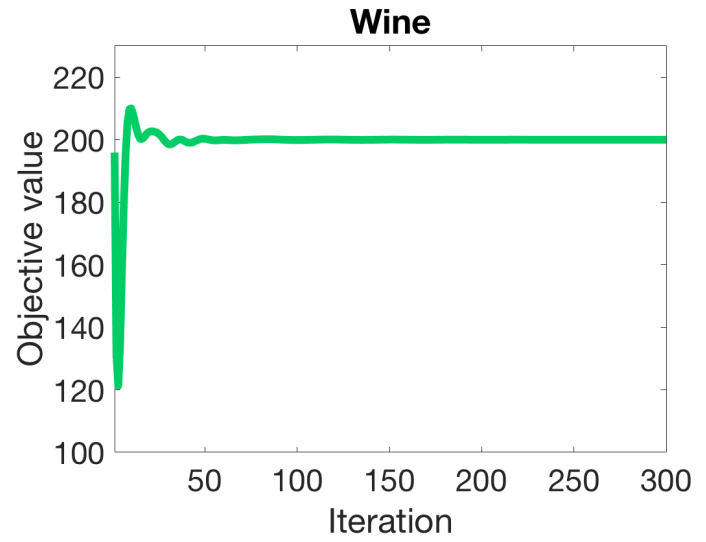}
\caption{Convergence of the K-means objective value using
FCKm-Mix with a random initialization on Wine dataset.}\label{fig:conv_3_appendix}
\end{center}
\end{figure}

\section{Auxiliary Results} We present some additional experimental results that augment the discussion made in the manuscript. Primarily, we provide empirical evidence that our FCKm method and its constrained variants converge to binary solutions that satisfy different constraints (pairwise and cardinality).

\textbf{Convergence for Unconstrained Clustering.} In Figure \ref{fig:conv_1_appendix}, we plot the K-means objective value at each ADMM iteration for an unconstrained clustering task with three clusters. Note that the initialization to this problem is a random threeway clustering. The objective decreases monotonically (after the first 2-3 iterations) and converges to a minimum value in approximately 6000 iterations. The optimization is stable with no perturbations at the onset of convergence.

In Figure \ref{fig:conv_2_appendix}, we plot the cluster vector for each of the three clusters being optimized at different iterations $(1,500,2500$, and 6530), i.e. we plot the three pieces of the label vector $\mathrm{x}$. In the first iteration, the initial clustering is done randomly, so it is binary but it does not lead to a good objective. As ADMM progresses, the continuous solution $\mathrm{x}$ becomes more and more binary, until it converges to a feasible binary solution where the three clusters are completely disjoint.

\textbf{Convergence for Constrained Clustering.} In Figure \ref{fig:conv_3_appendix}, we plot the K-means objective value at each ADMM iteration for a constrained clustering task with three clusters. In this task, we enforce cardinality, must-link, and cannotlink constraints onto the optimization. The initialization is taken to be a random assignment between the three disjoint clusters. In this case, the objective tends to be monotonically increasing after the first few iterations. This might seem counter-intuitive, since we are trying to minimize the objective. However, it must be noted that the continuous solution vector $\mathbf{x}$ in each ADMM iteration tends not to be feasible with respect to the enforced constraints. In other words, these constraints are being enforces more and more as the ADMM process proceeds, which forces the tradeoff between objective and feasibility. But, similar to the unconstrained case, the variation in objective is smooth and no perturbations are exhibited when ADMM begins to converge.

Moreover, Figure 4 plots the solution vectors for each cluster at four different iterations $(1,15,25,200)$. A similar behavior to the unconstrained case is encountered, where disjoint binary solution vectors are converged to. However, the notable difference is that we also report the number of cardinality (CardV), must-link (MLV), and cannotlink (CLV) violations at each of these iterations. We see that these violations gradually decrease until convergence occurs, when no violations persist.
\newpage


\EOD
\bibliography{cite}{}

\begin{thebibliography}{10}
\providecommand{\url}[1]{#1}
\csname url@samestyle\endcsname
\providecommand{\newblock}{\relax}
\providecommand{\bibinfo}[2]{#2}
\providecommand{\BIBentrySTDinterwordspacing}{\spaceskip=0pt\relax}
\providecommand{\BIBentryALTinterwordstretchfactor}{4}
\providecommand{\BIBentryALTinterwordspacing}{\spaceskip=\fontdimen2\font plus
\BIBentryALTinterwordstretchfactor\fontdimen3\font minus
  \fontdimen4\font\relax}
\providecommand{\BIBforeignlanguage}[2]{{%
\expandafter\ifx\csname l@#1\endcsname\relax
\typeout{** WARNING: IEEEtran.bst: No hyphenation pattern has been}%
\typeout{** loaded for the language `#1'. Using the pattern for}%
\typeout{** the default language instead.}%
\else
\language=\csname l@#1\endcsname
\fi
#2}}
\providecommand{\BIBdecl}{\relax}
\BIBdecl

\bibitem{karypis1999chameleon}
G.~Karypis, E.-H. Han, and V.~Kumar, ``Chameleon: Hierarchical clustering using
  dynamic modeling,'' \emph{Computer}, vol.~32, no.~8, pp. 68--75, 1999.

\bibitem{wu2011density}
B.~Wu and B.~Hu, ``Density and neighbor adaptive information theoretic
  clustering,'' in \emph{Neural Networks (IJCNN), The 2011 International Joint
  Conference on}.\hskip 1em plus 0.5em minus 0.4em\relax IEEE, 2011.

\bibitem{chapelle2005semi}
O.~Chapelle and A.~Zien, ``Semi-supervised classification by low density
  separation.'' in \emph{AISTATS}, 2005, pp. 57--64.

\bibitem{wagstaff2000clustering}
K.~Wagstaff and C.~Cardie, ``Clustering with instance-level constraints,''
  \emph{AAAI/IAAI}, 2000.

\bibitem{klein2002instance}
D.~Klein, S.~D. Kamvar, and C.~D. Manning, ``From instance-level constraints to
  space-level constraints: Making the most of prior knowledge in data
  clustering,'' Stanford, Tech. Rep., 2002.

\bibitem{ng2002spectral}
A.~Ng, M.~Jordan, and Y.~Weiss, ``On spectral clustering: Analysis and an
  algorithm,'' \emph{Advances in neural information processing systems},
  vol.~14, 2001.

\bibitem{bradley2000constrained}
P.~Bradley, K.~Bennett, and A.~Demiriz, ``Constrained k-means clustering,''
  \emph{Microsoft Research, Redmond}, 2000.

\bibitem{lu2010constrained}
Z.~Lu and H.~H. Ip, ``Constrained spectral clustering via exhaustive and
  efficient constraint propagation,'' in \emph{European Conference on Computer
  Vision}.\hskip 1em plus 0.5em minus 0.4em\relax Springer, 2010, pp. 1--14.

\bibitem{von2007tutorial}
U.~Von~Luxburg, ``A tutorial on spectral clustering,'' \emph{Statistics and
  computing}, 2007.

\bibitem{lu2007penalized}
Z.~Lu and T.~K. Leen, ``Penalized probabilistic clustering,'' \emph{Neural
  Computation}, 2007.

\bibitem{dai2003techniques}
B.-R. Dai, C.-R. Lin, M.-S. Chen \emph{et~al.}, ``On the techniques for data
  clustering with numerical constraints,'' \emph{Age}, 2003.

\bibitem{wu2013constrained}
B.~Wu, Y.~Zhang, B.-G. Hu, and Q.~Ji, ``Constrained clustering and its
  application to face clustering in videos,'' in \emph{Proceedings of the IEEE
  Conference on Computer Vision and Pattern Recognition}, 2013.

\bibitem{wagstaff2001constrained}
K.~Wagstaff, C.~Cardie, S.~Rogers, S.~Schr{\"o}dl \emph{et~al.}, ``Constrained
  k-means clustering with background knowledge,'' in \emph{ICML}, vol.~1, 2001,
  pp. 577--584.

\bibitem{hoppner2008clustering}
F.~Hoppner and F.~Klawonn, ``Clustering with size constraints,'' in
  \emph{Computational Intelligence Paradigms}.\hskip 1em plus 0.5em minus
  0.4em\relax Springer, 2008, pp. 167--180.

\bibitem{klawonn2006equi}
F.~Klawonn and F.~Hoppner, ``Equi-sized, homogeneous partitioning,'' in
  \emph{International Conference on Knowledge-Based and Intelligent Information
  and Engineering Systems}.\hskip 1em plus 0.5em minus 0.4em\relax Springer,
  2006.

\bibitem{shi2000normalized}
J.~Shi and J.~Malik, ``Normalized cuts and image segmentation,'' in \emph{IEEE
  Transactions on pattern analysis and machine intelligence}.\hskip 1em plus
  0.5em minus 0.4em\relax IEEE, 2000.

\bibitem{Wu2016ell_pBoxAA}
B.~Wu and B.~Ghanem, ``lp-box admm: A versatile framework for integer
  programming,'' \emph{IEEE transactions on pattern analysis and machine
  intelligence}, 2018.

\bibitem{COBWEB-1987}
D.~H. Fisher, ``Knowledge acquisition via incremental conceptual clustering,''
  \emph{Machine learning}, vol.~2, no.~2, pp. 139--172, 1987.

\bibitem{HMRF-constrained-2004}
S.~Basu, M.~Bilenko, and R.~J. Mooney, ``A probabilistic framework for
  semi-supervised clustering,'' in \emph{Proceedings of the tenth ACM SIGKDD
  international conference on Knowledge discovery and data mining}.\hskip 1em
  plus 0.5em minus 0.4em\relax ACM, 2004, pp. 59--68.

\bibitem{wu-iccv-2013}
B.~Wu, S.~Lyu, B.-G. Hu, and Q.~Ji, ``Simultaneous clustering and tracklet
  linking for multi-face tracking in videos,'' in \emph{Proceedings of the IEEE
  International Conference on Computer Vision}, 2013, pp. 2856--2863.

\bibitem{ratio-cut-2004}
I.~S. Dhillon, Y.~Guan, and B.~Kulis, \emph{A unified view of kernel k-means,
  spectral clustering and graph cuts}.\hskip 1em plus 0.5em minus 0.4em\relax
  Citeseer, 2004.

\bibitem{ding2015unified}
H.~Ding and J.~Xu, ``A unified framework for clustering constrained data
  without locality property,'' in \emph{Proceedings of the Twenty-Sixth Annual
  ACM-SIAM Symposium on Discrete Algorithms}.\hskip 1em plus 0.5em minus
  0.4em\relax SIAM, 2015.

\bibitem{duong2013declarative}
K.-C. Duong, C.~Vrain \emph{et~al.}, ``A declarative framework for constrained
  clustering,'' in \emph{Joint European Conference on Machine Learning and
  Knowledge Discovery in Databases}.\hskip 1em plus 0.5em minus 0.4em\relax
  Springer, 2013, pp. 419--434.

\bibitem{peng2007approximating}
J.~Peng and Y.~Wei, ``Approximating k-means-type clustering via semidefinite
  programming,'' \emph{SIAM journal on optimization}, 2007.

\bibitem{Hubert1985}
\BIBentryALTinterwordspacing
L.~Hubert and P.~Arabie, ``Comparing partitions,'' \emph{Journal of
  Classification}, vol.~2, no.~1, pp. 193--218, Dec 1985. [Online]. Available:
  \url{https://doi.org/10.1007/BF01908075}
\BIBentrySTDinterwordspacing

\bibitem{meilua2007comparing}
M.~Meila, ``Comparing clusterings--an information based distance,''
  \emph{Journal of multivariate analysis}, vol.~98, no.~5, pp. 873--895, 2007.

\bibitem{alqahtani2018deep}
A.~Alqahtani, X.~Xie, J.~Deng, and M.~Jones, ``A deep convolutional
  auto-encoder with embedded clustering,'' in \emph{Proceedings of the IEEE
  International Conference on Image Processing}, 2018, pp. 4058--4062.

\bibitem{alqahtani2019learning}
A.~Alqahtani, X.~Xie, J.~Deng, and M.~W. Jones, ``Learning discriminatory deep
  clustering models,'' in \emph{Proceedings of the International Conference on
  Computer Analysis of Images and Patterns}, 2019, pp. 224--233.

\bibitem{alqahtani2021deep}
A.~Alqahtani, M.~Ali, X.~Xie, and M.~W. Jones, ``Deep time-series clustering: A
  review,'' \emph{Electronics}, vol.~10, no.~23, p. 3001, 2021.

\end{thebibliography}
\bibliographystyle{IEEEtran}
\end{document}